\documentclass[10pt,twocolumn,letterpaper]{article}
\usepackage{cvpr}              % To produce the CAMERA-READY version
\usepackage{graphicx}
\usepackage{amsmath}
\usepackage{amssymb}
\usepackage{booktabs}
\usepackage{multirow}
\usepackage[table,xcdraw,dvipsnames]{xcolor}
\usepackage{makecell}
\usepackage{diagbox}
\usepackage{color}
\usepackage{bbding}
\usepackage{subcaption}
\usepackage[accsupp]{axessibility}
\usepackage[pagebackref,breaklinks,colorlinks]{hyperref}
\usepackage[capitalize]{cleveref}
\crefname{section}{Sec.}{Secs.}
\Crefname{section}{Section}{Sections}
\Crefname{table}{Table}{Tables}
\crefname{table}{Tab.}{Tabs.}

\def\cvprPaperID{9473} % *** Enter the CVPR Paper ID here
\def\confName{CVPR}
\def\confYear{2023}

\begin{document}

\title{Interventional Bag Multi-Instance Learning On Whole-Slide Pathological Images}

\author{Tiancheng Lin\textsuperscript{\rm 1,2} \qquad   Zhimiao Yu\textsuperscript{\rm 1,2}  \qquad  Hongyu Hu\textsuperscript{\rm 1,2} \qquad  Yi Xu\textsuperscript{\rm 1,2\thanks{Corresponding author.}} \qquad  Chang Wen Chen\textsuperscript{\rm 3}\\
\textsuperscript{\rm 1} Shanghai Key Lab of Digital Media Processing and Transmission, Shanghai Jiao Tong University\\
\textsuperscript{\rm 2} MoE Key Lab of Artificial Intelligence, AI Institute, Shanghai Jiao Tong University\\
\textsuperscript{\rm 3} The Hong Kong Polytechnic University, Hong Kong, China\\
{\tt\small \{ltc19940819, carboxy, mathewcrespo, xuyi\}@sjtu.edu.cn, changwen.chen@polyu.edu.hk}
% \author{Tiancheng Lin\textsuperscript{\rm 1}, Zhimiao Yu\textsuperscript{\rm 1}, Hongyu Hu\textsuperscript{\rm 1}, Yi Xu\textsuperscript{\rm 1*}, Chang Wen Chen\textsuperscript{\rm 2}\\
% \textsuperscript{\rm 1} Shanghai Jiao Tong University, Shanghai, China
% \textsuperscript{\rm 2} PolyU, Hong Kong, China\\
% {\tt\small \{ltc19940819, carboxy, mathewcrespo, xuyi\}@sjtu.edu.cn, changwen.chen@polyu.edu.hk}
% \and
% \\
}
\maketitle

%%%%%%%%% ABSTRACT
\begin{abstract}
Multi-instance learning (MIL) is an effective paradigm for whole-slide pathological images (WSIs) classification to handle the gigapixel resolution and slide-level label. 
Prevailing MIL methods primarily focus on improving the feature extractor and aggregator.
However, one deficiency of these methods is that the bag contextual prior may trick the model into capturing spurious correlations between bags and labels. This deficiency is a confounder that limits the performance of existing MIL methods. In this paper, we propose a novel scheme, \textbf{I}nterventional \textbf{B}ag \textbf{M}ulti-\textbf{I}nstance \textbf{L}earning (IBMIL), to achieve deconfounded bag-level prediction. Unlike traditional likelihood-based strategies, the proposed scheme is based on the backdoor adjustment to achieve the interventional training, thus is capable of suppressing the bias caused by the bag contextual prior. Note that the principle of IBMIL is orthogonal to existing  bag MIL methods. 
Therefore, IBMIL is able to bring consistent performance boosting to existing schemes, achieving  new state-of-the-art performance. Code is available at \url{https://github.com/HHHedo/IBMIL}.
\end{abstract}

%%%%%%%%% BODY TEXT
\section{Introduction}
\label{sec:intro}
\begin{figure}[t]
	\centering
	\begin{minipage}{1.0\linewidth}
% 		\centering
		\includegraphics[width=0.95\linewidth]{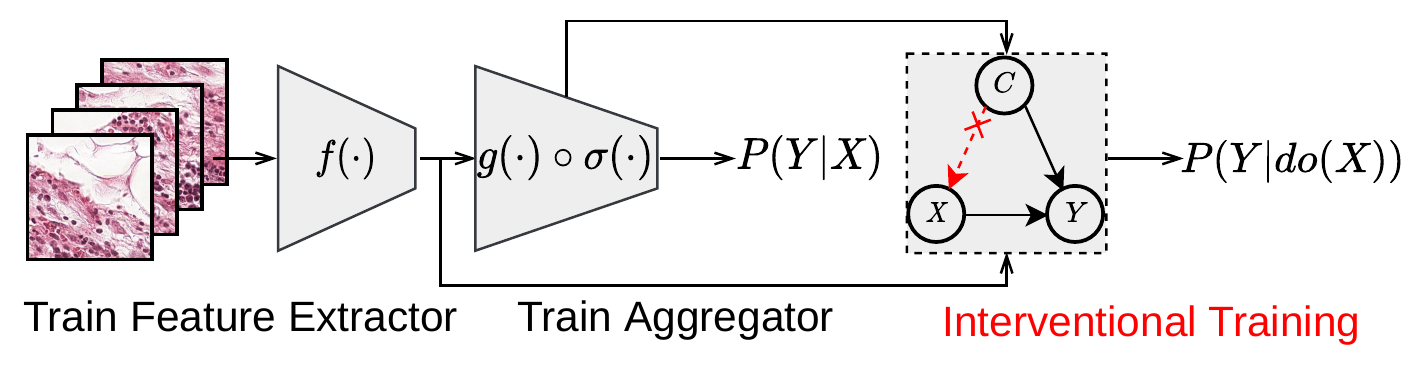}
		\subcaption{}
		\label{fig1a: pipeline}
	\end{minipage}
    \vspace{0.5cm}	
% 	\qquad

    \centering
    \begin{minipage}{1.0\linewidth}
	\begin{minipage}{0.49\linewidth}
		\centering
		
		\includegraphics[height=0.82\linewidth]{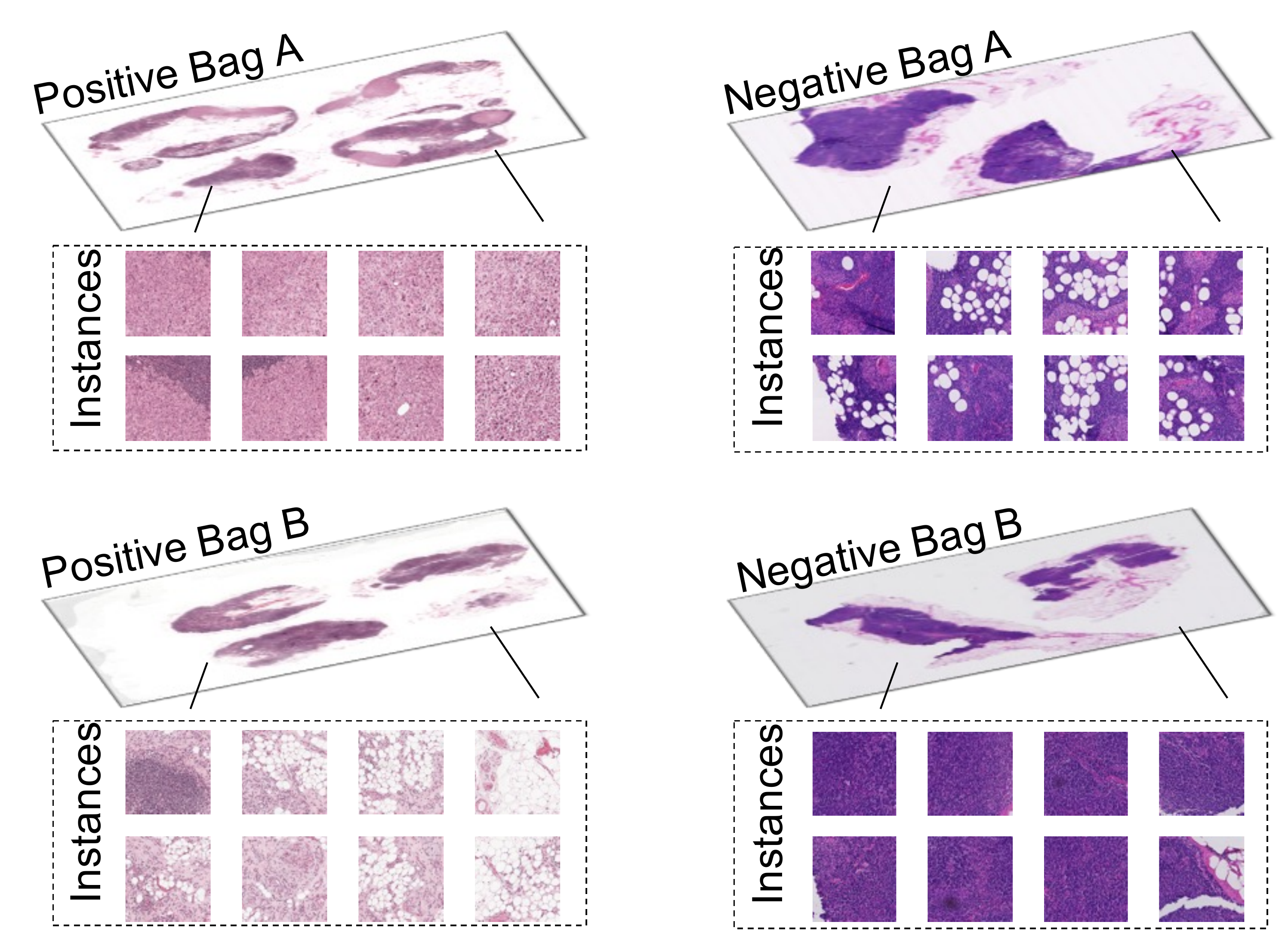}
		\subcaption{}
		\label{fig1b:WSI color}
	\end{minipage}
	\begin{minipage}{0.49\linewidth}
		\centering
		
		\includegraphics[height=0.82\linewidth]{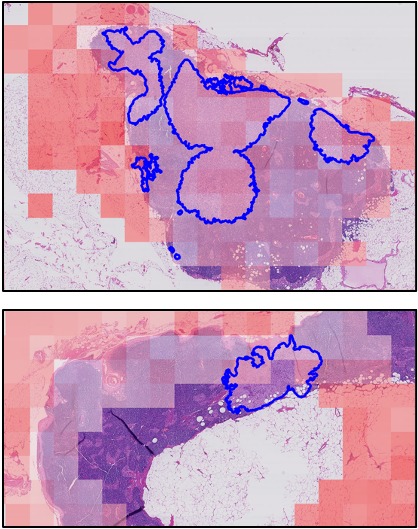}
		\subcaption{}
     \label{fig1c: attention}
	\end{minipage}
	\end{minipage}
\caption{
    (a) Traditional scheme and our interventional training.
    (b) Dataset bias. 
    (c) Unreasonable attention maps with right predictions.}
\end{figure}

The quantitative analysis of whole-slide pathological images (WSIs) is essential for both diagnostic and research purposes~\cite{review09}.
Beyond complex biological structures, WSIs are quite different from natural images in the gigapixel resolution and expensive annotation, which is thus formulated as a multi-instance learning (MIL)~\cite{dietterich1997MIL} problem: treating each WSI as a labeled bag and the corresponding patches as unlabeled instances.
Such a \emph{de facto} paradigm has been demonstrated in extensive tasks on WSIs, $e.g.$, classification~\cite{hou2016cvpr, mutation, MSI, VAE+GAN+GCN}, regression~\cite{zhu2017,DeepMISL,DeepAttnMISL} and segmentation~\cite{camel}.
The prevailing scheme for WSI classification --- bag-level MIL --- is depicted in~\cref{fig1a: pipeline}. Given the patchified images as instances, each instance is embedded in vectors by a feature extractor in the first stage. Second, for each bag, their corresponding instance features are aggregated as a bag-level feature for classification.

More and more new frameworks are proposed to improve the two stages following this scheme~\cite{NIC,li2021dual,shao2021transmil, zhang2022dtfd}.
It is convinced that learning better instance features and modeling more accurate instance relationships can bring better performance of MIL.
While we have witnessed the great efforts, they still leave the ``bag contextual prior" issue unsolved: the information shared by bags of the same class but irrelevant to the label, which may affect the final predictions. 
For example, in~\cref{fig1b:WSI color}, due to the dataset bias, most of the instances in the positive bags are stained pink but purple in the negative bags. 
The co-occurrence of specific color patterns and labels may mislead the model to classify bags by color statistics instead of the key instances --- the more pink instances a bag contains, the more likely it is a positive bag. ~\cref{fig1c: attention} illustrates another example: even if the prediction is correct, the underlying visual attention is not reasonable, where the high attention scores are put on the disease-irrelevant instances outside the blue curves in the bags.
From the causal lens, the bag contextual prior is a confounder that opens up a backdoor path for bags and labels, causing spurious correlations between them.
To suppress such a bias, we need a more efficient mechanism for the actual causality between bags and labels, $i.e.$, the bag prediction is based on the bag's content ($e.g.$, key instances), which can not be fully achieved only by above mentioned new frameworks.

% to empowering the existing state-of-the-art (SoTA) methods with the deconfounded training strategy
In fact, it is challenging to achieve unbiased bag predictions as such a bias happens in the  data generation -- the tissue preparations, staining protocols, digital scanners, $etc$.
In this paper, we propose a novel MIL scheme, \textbf{I}nterventional \textbf{B}ag \textbf{M}ulti-\textbf{I}nstance \textbf{L}earning (IBMIL), to tackle this challenge.
In particular, we propose a structure causal model (SCM)~\cite{pearl2016causal} to analyze the causalities among bag contextual prior, bags and labels. 
The key difference of IBMIL is that it contains another stage of interventional training (see ~\cref{fig1a: pipeline} right). 
Given the  aggregator trained in the second stage, instead of directly using it for inference via \textbf{likelihood:} $P(Y|X)$, we apply it for the approximation of confounders. 
With the confounders observed, we eliminate their effect via the backdoor adjustment formulation~\cite{backdoor}, where the intuitive understanding is: if a  WSI model can learn from   ``purple'' and ``pink'' positive/negative bags, respectively, then the bag context of color will no longer confound the recognition. 
Therefore, our IBMIL is fundamentally different from the existing scheme as we use a \textbf{causal intervention:} $P(Y|do(X))$  for bag prediction.

We conduct experiments on two public WSI datasets, $i.e.$, Camelyon16~\cite{CAMELYON16} and TCGA-NSCLC. 
Experimental results show that IBMIL is agnostic to both feature extractors  and aggregation networks, $i.e.$, it brings consistent performance boosting to all compared state-of-the-art MIL methods in the WSI classification tasks. 
Further ablation  studies and analyses demonstrate the effectiveness of interventional training.

\section{Related Work}
% \subsection{Pathological Multi-Instance Learning}
\subsection{Instance-level MIL on WSIs}
Instance-level MIL represents each instance by a score and aggregates instance scores into a bag score. One widely used baseline is SimpleMIL ~\cite{SimpleMIL,NM2018}, which directly propagates the bag label to its instances.
When applying SimpleMIL for WSIs, the unbalanced dataset could result in noisy instance-level supervision since a WSI ($e.g.$, Camelyon16) might only contain a small portion of a disease-positive tissue in clinic~\cite{li2021dual}.
% It has been demonstrated to be quite effective but suffers from the noisy instance-level supervision, which can be a serve problem when applying it to WSIs. 
The following works in this line improve this baseline via various modifications. 
\textit{Cleaner annotations}: SemiMIL~\cite{wang2019weakly} directly introduces cleaner annotations for partial instances with the help of pathologists, where these annotated regions are assigned with larger weights as they carry higher confidence. 
\textit{Instance selection}: PatchCNN~\cite{hou2016cvpr} selects instances via a delicate thresholding scheme at both WSI and class levels. Similarly, Top-$k$ MIL~\cite{chikontwe2020topk} only uses the top-$k$ instances for each bag, but the fixed number of selected instances fails to make a trade-off between preserving clean instances and discarding noisy instances. RCEMIL~\cite{RCE} proposes rectified cross-entropy (RCE) loss to select instances in a softer manner, while the loss requires statistics of possible abnormal tissues among all WSIs. 
% robust loss~\cite{RCE}, $etc$.
% Some works try to only use the clean instances for training. 
% For example, , while Chikontwe et al.
% Similarly, 
More recently, IMIL~\cite{lin2022interventional} summarizes the previous works from a causal lens and propose IMIL to select instance via causal intervention and effect.
However, the performance of instance-level MIL methods is usually inferior to bag-level counterparts~\cite{wang2018revisiting}.
% More recently, Chikontwe et al. \cite{chikontwe2020topk} employs top-\textit{k} instance selection \cite{wen2016centerloss}, but the fixed number of selected instances fails to make a trade-off between preserving clean instances and discarding noisy instances.
% Indeed, all these methods tackle the noisy supervision by selecting and re-weighting instances. 

\subsection{Bag-level MIL on WSIs} 
The instances are represented as embedding vectors and classified by bag-to-bag distance/similarity or a bag classifier \cite{wang2018revisiting}. 
% Many works focus on designing novel distance metrics and aggregation operators, such as bag similarity network \cite{wang2019BSN}, dynamic pooling \cite{yan2018dynamic}, attention mechanism \cite{AMIL, li2021dual}, recurrent neural network \cite{clinical-grade}, graph convolutional network \cite{VAE+GAN+GCN} and so on.
Conducting bag-level MIL on WSI is non-trivial, because the intermediate results of all patches still need to be stored in memory for backpropagation. 
Therefore, some recently proposed frameworks separate the training of instance-level feature extractors and aggregation networks, resulting in a two-stage modeling approach~\cite{NIC,li2021dual,shao2021transmil, zhang2022dtfd}.
They contribute differently at both stages. 
For the feature extractor, they introduce different \textit{architectures} from convolutional neural networks (CNNs) to transformer-based models~\cite{transformerbbmil}, and \textit{training paradigms} from ImageNet pre-training~\cite{DeepMISL,DeepAttnMISL,CNN+GBDT} to self-supervised learning~\cite{VAE+GAN+GCN,NIC,CPC+MIL}. 
% It is naturally believed that good instance-level features can guarantee the MIL performance.
Simultaneously, many works pay attention to new  designs of aggregation networks, from non-parametric poolings, $e.g.$, max/mean-poolings~\cite{wang2018revisiting}, to learnable ones,  $e.g.$, graph convolution networks~\cite{VAE+GAN+GCN} and attention mechanisms~\cite{AMIL,li2021dual, shao2021transmil,zhang2022dtfd}. 
Our work lies in this line but aims at empowering these existing methods. Thus the contributions are orthogonal.
% Moreover, rarely did they explicitly explore the MIL assumption.
% Our work lies in this line and aims at accessing the key factors in MIL for WSIs.

\subsection{Causal Inference in Computer Vision}
Causal inference, a general framework, has been introduced to various computer vision tasks, including classification~\cite{CIL,tang2020long,yue2020interventional}, semantic segmentation~\cite{zhang2020causal,causalcam}, unsupervised representation learning~\cite{wang2020visual,mitrovic2020representation,relic2}  and so on. 
In MIL problems, StableMIL~\cite{zhangstable} takes ``adding an instance to a bag'' as a treatment for bag-level prediction, while IMIL~\cite{lin2022interventional} uses inverse probability weighting and causal effects for instance-level tasks. 
Unlike them, our IBMIL is based on backdoor adjustment formulation and works as a general framework to empower existing bag-level MIL for WSI classification tasks.
% and explore the bag-level relations for 

\section{Method}

\subsection{Preliminaries}
\noindent\textbf{MIL formulation.} Due to the gigapixel resolution and lacking fine-grained labels, performing downstream tasks on WSIs is formulated as an MIL problem, such that each WSI is treated as a \textit{labeled bag} with corresponding patches as \textit{unlabeled instances}. 
Take binary classification as an example, let $X = \{(x_1,y_1),...,(x_n, y_n)\}$ be a WSI bag, which contains $n$ instances of $x_i$. The instance-level labels $\{y_i ,...,y_n\}$  are unavailable.
Under the standard MIL assumption, the bag label $Y$ is further given by:
\begin{equation}
Y= \begin{cases}0, & \text { iff } \sum y_i=0 \\ 1, & \text { otherwise } \end{cases}
\end{equation}
which can be modelled by max-pooling~\cite{hou2016cvpr}. 
A general three-stage approach goes like 1) \textit{Instance transformation:} a feature extractor $f(\cdot)$ is trained for instance-wise features $b$, 2) \textit{Instance combination:} the pooling operation $\sigma(\cdot)$ is targeted for bag feature $B$, 3) \textit{Bag transformation:}  a downstream classifier $g(\cdot)$ is used for prediction, which can be formulated as:
%In particular, a feature extractor $f$ and a pooling operation  is applied to represent bag feature $B$, followed by a downstream classifier $g$ to predict the label of $X$, which is formulated as:
\begin{equation}
    b_i = f(x_i), B = \sigma(b_1,\cdots,b_n), \hat{Y} = g(B),
\end{equation}
where the pooling $\sigma(\cdot)$ should be a permutation-invariant function~\cite{AMIL} for the spatial-invariant MIL method. Some works further absorb the classifier $g(\cdot)$ into the pooling operation $\sigma(\cdot)$, referred to as aggregator/aggregation networks.
When applying the MIL methods for WSIs, it should be noted that 1) the diagnosis for WSI analysis can be based on different tissue regions with multiple concepts --- the collective MIL assumption, 2)  the bag length $n$ for a WSI can be extremely large, $e.g.$, about 8,000 on average\cite{li2021dual}.
Therefore, the bag MIL methods for WSIs are with \textit{learnable aggregators} and trained in a \textit{two-stage} procedure, $i.e.$, training the feature extractor and aggregator stage by stage. 
% In the first stage, the feature extractor  is trained \textcolor{red}{in a weakly supervised  paradigm~\cite{} or self-supervised  paradigm~\cite{}}. 
% In the second stage,  the aggregator is trained upon frozen feature extractor (or extracted features).
Current works mainly follow this formulation and improve the framework from both feature extractor and aggregator, while our proposed method aims to empower existing works from a causal perspective.

\begin{figure}[t]
  \centering
%   \includegraphics[width=0.6\columnwidth]{figs/causal.drawio.pdf}
%   \caption{An illustration of causal graph for bag MIL framework. }
  \label{fig:causal}
%   \vspace{-0.3cm}
    \begin{subfigure}[b]{0.4\linewidth}
    \centering
     \includegraphics[height=3cm]{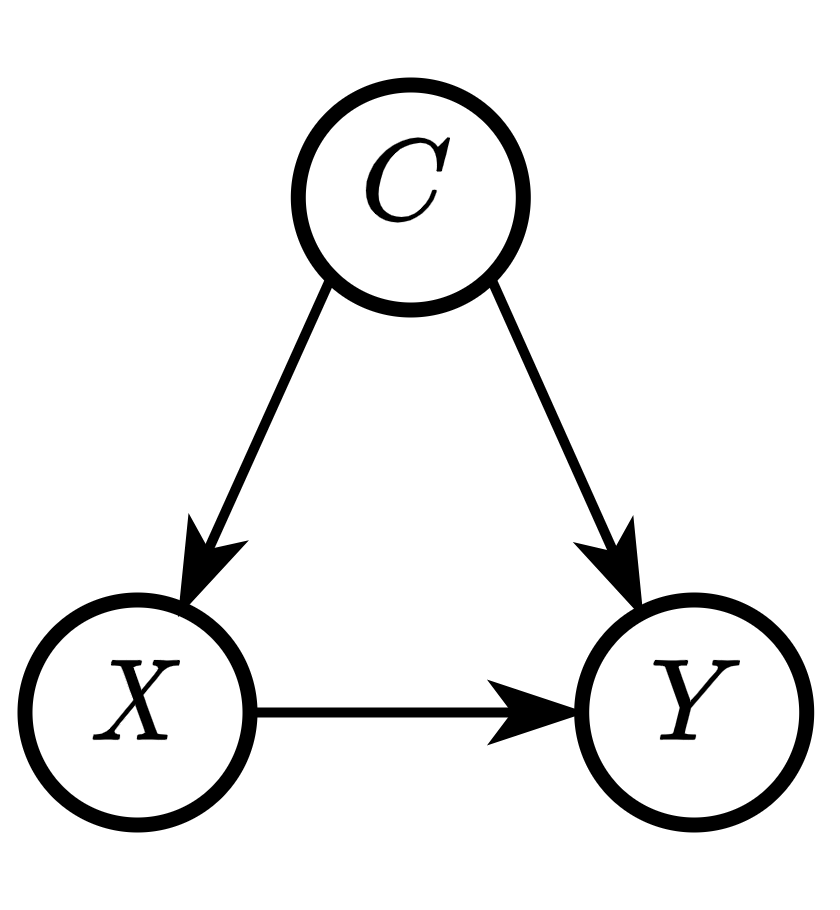}
     \caption{Causal graph}
     \label{fig:Causal graph}
    \end{subfigure}
    % \hfill
    \begin{subfigure}[b]{0.4\linewidth}
    \centering
     \includegraphics[height=3cm]{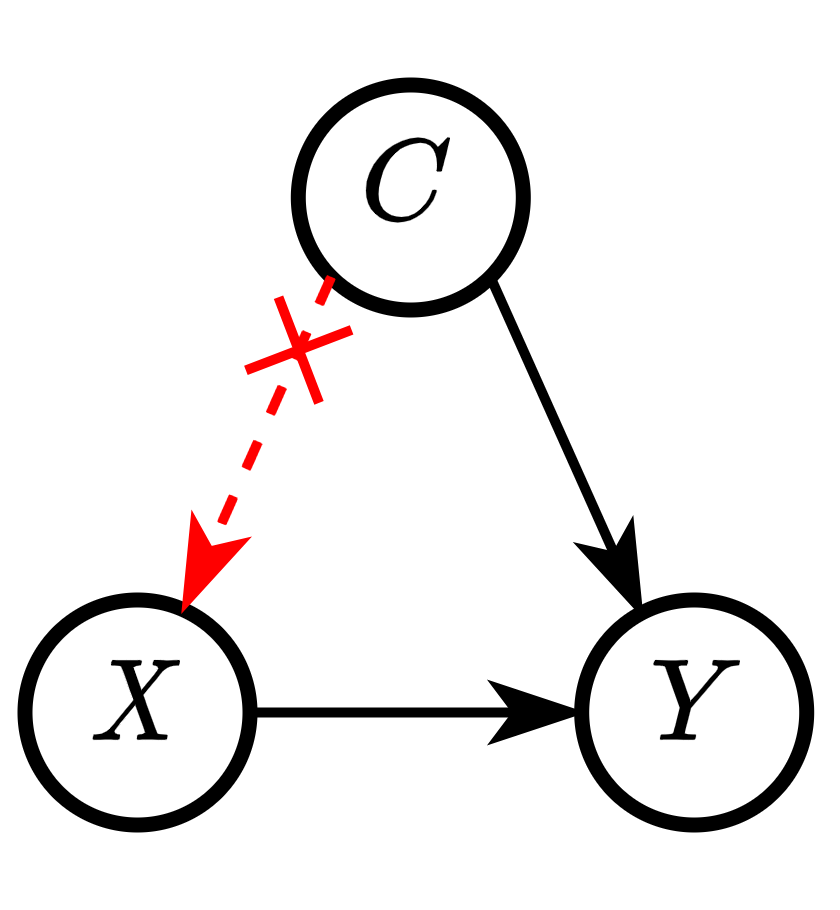}
     \caption{$P(Y|do(X))$}
     \label{fig: do}
    \end{subfigure}
    \caption{An illustration of causal graph for bag MIL framework.}
    \label{fig1:stati}
\end{figure}

\noindent\textbf{Analysis MIL through causal inference.}
As shown in~\cref{fig:Causal graph}, we formulate the MIL framework as a causal graph (\textit{a.k.a}, Pearl's structural causal model or SCM~\cite{pearl2016causal}), which contains three nodes: $X$: whole-slide pathological image (bag), $Y$: bag label, $C$: bag contextual information.

\textbf{$X\rightarrow Y$:} This path indicates that the MIL model can learn to predict the bag label on the bag content, $e.g.$, key instances.

\textbf{$C\rightarrow X$:} This path indicates the generation of the whole-slide pathological image. Due to the differences in tissue preparations, staining protocols and digital scanners, the appearance of WSIs can be significantly affected, potentially introducing biases.

\textbf{$C\rightarrow Y$:} This path indicates that the bag prediction is affected by the contextual prior information in the training dataset. For example, in~\cref{fig1b:WSI color}, an MIL model predicts all bags with purple color as positive regardless of content information related to the real label.

In the causal graph, $C$ confounds $X$ and $Y$ via the backdoor path $X \rightarrow C \rightarrow Y$ and causes a spurious correlation between them, which prevents learning robust bag MIL models. For example, the model may wrongly predict the bags when the data are out-of-distribution, $i.e.$, with different context prior.
% $i.e.,$ learning to predict the bag label only based on. The existence of back-door 
An ideal MIL method should capture the true causality between $X$ and $Y$, but the conventional correlation of $P(Y|X)$ fails to do so, as such a spurious correlation is inevitable.
Therefore, we instead seek to use the causal intervention $P(Y|do(X))$, where the do-operation $do(\cdot)$ means forcibly assigning a specific value to the variable $X$. As shown in~\cref{fig: do}, it can be considered as a modification of the graph ---  cutting off the backdoor path, thus mitigating the bias caused by confounders. The ideal way of $do(\cdot)$ is the random controlled trials~\cite{pearl2018book} --- enumerating each bag with all possible contexts, which is impossible in practice. Next, we propose a practical intervention method to remove the confounding effect caused by the bag contextual prior. 
\subsection{Interventional Bag Multi-Instance Learning}
\begin{figure*}[t]
\normalsize
  \centering
  \includegraphics[width=\textwidth]{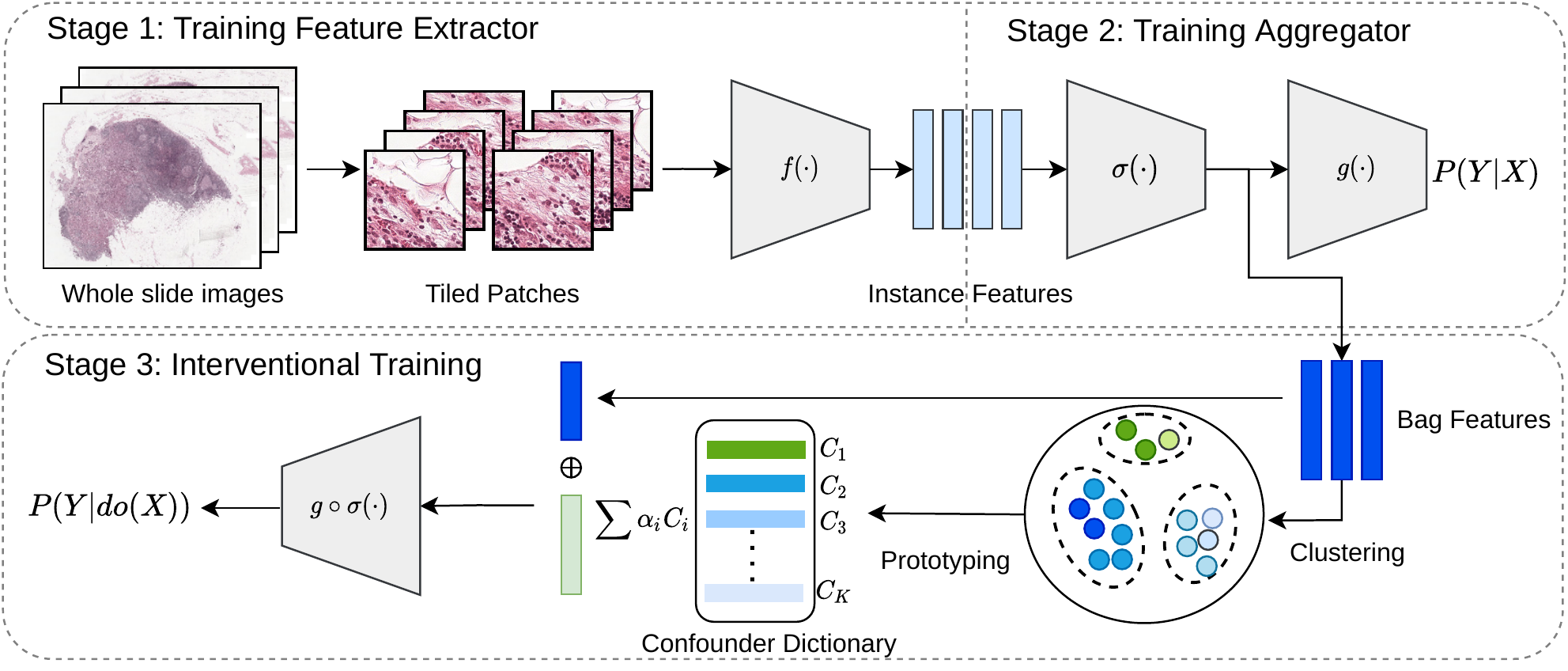}
  \caption{Overview of our proposed Interventional Bag Multi-Instance Learning (IBMIL). Our contribution is to introduce interventional training to the traditional two-stage scheme.}
  \label{fig:main_fig_framework}
%   \vspace{-0.5cm}
\end{figure*}
We propose to use the backdoor adjustment formulation to achieve the causal intervention: $P(Y|do(X))$  for bag-level prediction.
Formally, we have the backdoor adjustment for the graph in~\cref{fig:Causal graph} as:
% \begin{equation}
% P(Y \mid d o(X))=\sum\nolimits_{i=1}^k  P(Y \mid X, h(X,c_i)) P(c_i),
% \label{eq:backdoor}
% \end{equation}

\begin{equation}
P(Y \mid d o(X))=\sum _{i}  P(Y \mid X, h(X,c_i)) P(c_i),
\label{eq:backdoor}
\end{equation}
%where $k$ is the number of confounders 
where $c_i$ loops over the confounder set and $h(\cdot)$ is a function defined later in \cref{eq:softmax}. Different from Bayes rule,  in \cref{eq:backdoor}, $c_i$ is no longer affected by $X$ but subject to its prior $P(c_i)$, since the causal intervention forces  $X$ to incorporate each $c_i$ fairly. 
Now, we are ready to introduce our interventional bag multi-instance learning stage by stage. \cref{fig:main_fig_framework} illustrates the overview of IBMIL.
% assume confouders
% formulation, definitions. difference from Bayes rule, no observational bias, independent, incorporate each c fairly, subject to a prior
% how to form the IBMIL, three steps 

\noindent\textbf{Stage 1: Training feature extractors.}
We learn a feature extractor $f(\cdot)$ on the patchified images of WSIs $\{x_1 ,...,x_n\}$, aiming at encoding each instance as a discriminative feature vector. 
% In the deep learning era, it can be a CNN, transformer or hybrid network, and it can be trained by different training paradigms for representation learning.
% Here we focus on the pre-training paradigms under the same architecture. 

%  of different feature extractor are given in section x and appendix x.

\noindent\textbf{Stage 2: Training aggregators.}
Given the features of instances  $\{b_1 ,...,b_n\}$, the aggregator employs MIL pooling $\sigma(\cdot)$  to assemble them into a bag feature $B$ sequentially or simultaneously, and a classifier $g(\cdot)$  for discrimination. Formally,  the loss for training aggregator is defined as:
\begin{equation}
\mathcal{L}=-\frac{1}{N} \sum_{i=1}^N Y_i \log \hat{Y}_i+\left(1-Y_i\right) \log \left(1-\hat{Y}_i\right),
\end{equation}
where $N$ is number of bags in the training set. 
Note our IBMIL is no not limited to specific feature extractor or aggregators, including the architectures and training paradigms. Please refer to \cref{sec:exp}  for our choices.

\noindent\textbf{Stage 3: Causal intervention via backdoor adjustment.}
The traditional two-stage bag MIL stops at stage 2 and uses the trained models for inference directly.
Instead, we introduce another stage of interventional training, which needs the practical implementation of \cref{eq:backdoor}. 
Note that backdoor adjustment assumes that we can observe and stratify the confounders of a bag context. Thanks to the powerful ability of deep MIL models, context information is naturally encoded in the higher-level layers~\cite{zhang2020causal,lin2022interventional}. 
% First, we use a confounder dictionary $C=\left[c_1, \ldots, c_K\right]$ for approximation to constitute the confounder set, as collecting all confounders is impossible.
To constitute the confounder set, we use a confounder dictionary
%$\{\boldsymbol{c_k} \}_{k=1}^K$
%$\boldsymbol{C}=\left[\boldsymbol{c}_1, \ldots, \boldsymbol{c}_K\right]$ 
$C=\left[c_1, \ldots, c_K\right]$
for approximation, as collecting all confounders is impossible. 
Given the trained feature extractor and aggregator, we use $K$-means over all the bag features in the training set, partitioning the bags into clusters. 
We average the bag features of each cluster to represent a confounder stratum $c_i$, resulting in a  confounder dictionary with the shape of $d \times K$, where $d$ is the dimension of bag features.
Note that our approximation is reasonable in that these global clusters are susceptible to the visual biases~\cite{sharma2021cluster512}, which is exactly the confounders.
Then, we define:
% \begin{equation}
% %   \begin{array}{l}
%     h(\boldsymbol{X},\boldsymbol{c})=\sum_{i=1}^k \alpha_i \boldsymbol{c}_i,\\ 
%     \alpha_i=\operatorname{softmax}\left(\frac{\left(\mathbf{W}_1 \boldsymbol{B}\right)^T\left(\mathbf{W}_2 \boldsymbol{c}_i\right)}{\sqrt{l}}\right),
% %   \end{array}
% \label{eq:softmax}
% \end{equation}
\begin{equation}
  \begin{array}{l}
    h(X,c_i)= \alpha_i c_i,\\ 
    	\left[\alpha_1,\cdots ,\alpha_K\right]=\operatorname{softmax}\left(\frac{\left({W}_1 B\right)^T\left({W}_2 C\right)}{\sqrt{l}}\right),
  \end{array}
\label{eq:softmax}
\end{equation}
where $B=\sigma(f(X))$ is the bag feature, ${W}_1, {W}_2 \in \mathbb{R}^{l \times d}$ are two learnable projection matrices to project bag feature ${B}$ and confounder ${C}$ into a joint space, and $\sqrt{l}$ is used for feature normalization~\cite{wang2020visual}. Since the prediction comes from both bag $X$ and confounder $C$ (see~\cref{fig:Causal graph}), we further define
\begin{equation}
    P(Y \mid X, h(X,c_i)) = P(Y \mid B \oplus h(X,c_i)),
    \label{eq:con}
\end{equation}
where $\oplus$ denotes vector concatenation, and other implementations can be found in ablation studies.
We assume $P({c}_i)$ is a uniform prior of $1/K$ for a safe estimation, and a more reasonable assumption, $e.g.$, incorporating expert knowledge, will be our future work. 
Plugging \cref{eq:softmax}, \cref{eq:con} and defined $P({c}_i)$ into \cref{eq:backdoor}, we are ready to calculate $P(Y|do(X))$ via passing the network multiple times. In practice, to avoid the expensive cost, we further apply Normalized Weighted Geometric Mean~\cite{NWGM} to move the outer sum into the Softmax:
% \begin{equation}
% \begin{aligned}
% &P(Y \mid d o(X))  \\
% &\approx P(Y \mid B \oplus  \sum_{i=1}^K \operatorname{softmax}\left(\frac{\left({W}_1 {B}\right)^T\left({W}_2 {c}_i\right)}{\sqrt{l}}\right) {c}_i P(c_i) )  .
% \label{eq:final}
% \end{aligned}
% \end{equation}
\begin{equation}
\begin{aligned}
P(Y \mid d o(X)) \approx P\left(Y \mid B \oplus  \sum\nolimits_{i=1}^K \alpha_i {c}_i P(c_i) \right)  .
\label{eq:final}
\end{aligned}
\end{equation}
Thus,  backdoor adjustment can be achieved by one feed-forward of the network. 
% The derivation is given in appendix. 
\subsection{Justification}
In our implementation of the causal intervention, there are some aspects we need to discuss further. 

\noindent\textbf{Compatible with large-scale unlabelled datasets.} 
We constitute the confounder set in an unsupervised fashion.
One alternative implementation is to use the available bag labels for guidance, preserving the intra-class variation and capturing the class-relevant characteristics .
There are two main reasons  for our choice. 1) The unsupervised fashion makes our scheme compatible with large-scale unlabelled datasets, $e.g.$, The Cancer Genome Atlas (TCGA), for better approximation of confounders. 2) The confounder could be irrelevant to the class identity, $e.g.$, the stain color of positive and negative instances can be the same. 
We explore the other implementations in~\cref{sec:dis}.

\noindent\textbf{One possible more elegant scheme.} As we need the trained aggregator to generate the bag features (the stage 2), one more stage is needed to retrain the aggregator (the stage 3).
We are thus motivated to further simplify our scheme to avoid extra computational cost.
Specifically, we can achieve the bag features by applying the traditional non-parametric aggregators, $e.g.$, max/mean-pooling, to the instances in a bag. 
It is inspired by the fact that these non-parametric aggregators serve as strong baselines, and we conjecture that statistic bag information they provide can be used for a reasonable approximation of confounders.
Therefore, we can omit the stage 2. 
The experiment results in \cref{sec:dis}  support that our scheme can be more elegant.
% Though we will prove the improvement does not come from longer training (\textcolor{red}{see section x})

\noindent\textbf{Connection to other methods.} \textit{Embedding-based MIL:} As we approximate the confounder set based on bag features, these confounders can be seen as a denoised abstraction of bag features. From this perspective, we share the same spirit with the embedding-based MIL~\cite{wang2018revisiting}, $i.e.$, exploring the relations between bags. That means our IBMIL also explains the effectiveness of embedding-based MIL.
\textit{Color Normalization}: Some works~\cite{colornor} propose color normalization methods for H\&E stained WSIs. However, color is just one of the confounders, and some confounders are even unobserved. Our method does not focus on color only, and thus is the more reasonable partially observed children of the unobserved confounder~\cite{confchild}; \textit{Instance augmentation}:  IMIL~\cite{lin2022interventional} uses strong instance augmentation to train the feature extractor for instance prediction. However, the augmentation may affect the statistical information in the bag. Therefore, our method is more suitable for bag MIL. Remix~\cite{remix} proposes  data augmentations for MIL by exploring the relations of instances, but our method explores the bag-level relations based on the causal theory.
%def of bag information

\section{Experiments}
\label{sec:exp}

\begin{table*}[t]
\caption{Main results (\%) on Camelyon16 and TCGA-NSCLC. }
\centering
\resizebox{0.87\textwidth}{!}
{\begin{tabular}{clccccccccc} 
\toprule
\multicolumn{3}{c}{\multirow{2}{*}{\diagbox[width=12em]{\textbf{Method}}{\textbf{Performance}}}} & \multicolumn{4}{c}{Camelyon16} & \multicolumn{4}{c}{TCGA-NSCLC} \\ 
\cmidrule(lr){4-7}  
\cmidrule(lr){8-11} 
\multicolumn{3}{r}{} & Precision & Recall  & Accuracy & AUC & Precision & Recall  & Accuracy & AUC\\
\hline
\multirow{12}{*}{\rotatebox{90}{\makecell{ResNet-18 \\ImageNet pretrained}}}
&\multirow{3}{*}{ABMIL}&&86.71&	81.71&	84.50&	84.07& 82.75&	85.84&	81.43&	88.95 \\
&&+IBMIL                &88.58&	87.14&	88.37&	90.43 & 85.42&	85.17&	85.24&	91.26 \\ 
&&${\Delta}$ & \textcolor{ForestGreen}{+1.87} & \textcolor{ForestGreen}{+5.43} & \textcolor{ForestGreen}{+3.87} & \textcolor{ForestGreen}{+6.36} & \textcolor{ForestGreen}{+2.67}& \textcolor{gray}{-0.67}& 
\textcolor{ForestGreen}{+3.81}
& \textcolor{ForestGreen}{+2.31}\\ \cline{2-11}  
&\multirow{3}{*}{DSMIL}&&84.56&	82.95&	84.50&	87.16& 
80.56&	85.78&	77.62&	86.88\\
&&+IBMIL 
&90.17&	86.20&	88.37&	87.69& 
81.98&	86.25&	80.00&	87.19 \\ 
&&${\Delta}$ & \textcolor{ForestGreen}{+5.61}& \textcolor{ForestGreen}{+3.25}
& \textcolor{ForestGreen}{+3.87}
& \textcolor{ForestGreen}{+0.53}
& \textcolor{ForestGreen}{+1.42}
& \textcolor{ForestGreen}{+0.47}
& \textcolor{ForestGreen}{+2.38}
& \textcolor{ForestGreen}{+0.31}\\ \cline{2-11}  
&\multirow{3}{*}{TransMIL}&& 85.43	&81.06&	83.72&	81.29 &85.46	&85.31&	85.24&	90.70\\

&&+IBMIL & 83.14	&82.93&	83.72&	88.71 &85.80	&87.06	&85.24	&92.54\\ 
&&${\Delta}$ & \textcolor{gray}{-2.29}
& \textcolor{ForestGreen}{+1.87}
& \textcolor{gray}{0.00}
& \textcolor{ForestGreen}{+7.42}
& \textcolor{ForestGreen}{+0.34}
& \textcolor{ForestGreen}{+1.75}
& \textcolor{gray}{0.00}
& \textcolor{ForestGreen}{+1.84}
\\ \cline{2-11}  
&\multirow{3}{*}{\makecell{DTFD-MIL \\(MaxS) } }&& 84.85&	80.09&	82.95&	82.77&
82.29&	83.77&	81.90&	88.91\\

&&+IBMIL & 89.53&	86.51&	88.37&	89.51&
83.25&	82.96&	82.86&	90.50\\

&&${\Delta}$& \textcolor{ForestGreen}{+4.68}
& \textcolor{ForestGreen}{+6.42}
& \textcolor{ForestGreen}{+5.42}
& \textcolor{ForestGreen}{+6.74}
& \textcolor{ForestGreen}{+0.96}
& \textcolor{gray}{-0.81}
& \textcolor{ForestGreen}{+0.96}
& \textcolor{ForestGreen}{+1.59}
\\ 
\cline{2-11}  

% ####################
\hline
\multirow{12}{*}{\rotatebox{90}{\makecell{CTransPath \\ SRCL}}}
&\multirow{3}{*}{ABMIL}&&91.84&	89.09&	90.70&	92.33
&90.76	&90.40&	90.48&	95.87 \\

&&+IBMIL& 95.29&	92.31&	93.80&	93.83 & 91.92&	91.93&	91.90&	96.91 \\ 
&&${\Delta}$
& \textcolor{ForestGreen}{+3.45}
& \textcolor{ForestGreen}{+3.22}
& \textcolor{ForestGreen}{+3.10}
& \textcolor{ForestGreen}{+1.50}
& \textcolor{ForestGreen}{+1.16}
& \textcolor{ForestGreen}{+1.53}
& \textcolor{ForestGreen}{+1.42}
& \textcolor{ForestGreen}{+1.04}\\ 
\cline{2-11}  
&\multirow{3}{*}{DSMIL}&& 89.24&	88.10&	89.15&	93.26&
92.11&	92.79&	90.95&	97.13\\
&&+IBMIL & 91.05&	91.30&	91.47&	95.20 
&92.05&	93.82&	91.43&	97.51 \\ 
&&${\Delta}$
& \textcolor{ForestGreen}{+1.81}
& \textcolor{ForestGreen}{+3.20}
& \textcolor{ForestGreen}{+2.31}
& \textcolor{ForestGreen}{+1.94}
& \textcolor{gray}{-0.06}
& \textcolor{ForestGreen}{+1.03}
& \textcolor{ForestGreen}{+0.48}
& \textcolor{ForestGreen}{+0.38}
\\ 
\cline{2-11}  
&\multirow{3}{*}{TransMIL}&& 95.83&	93.27&	94.57&	95.88 &92.05&	93.82&	91.90&	95.55\\

&&+IBMIL &96.48&	95.50&	96.12&	97.00 &93.94&	93.76&	93.81&	97.24\\ 
&&${\Delta}$
& \textcolor{ForestGreen}{+0.35}
& \textcolor{ForestGreen}{+2.33}
& \textcolor{ForestGreen}{+1.55}
& \textcolor{ForestGreen}{+1.12}
& \textcolor{ForestGreen}{+1.89}
& \textcolor{gray}{-0.06}
& \textcolor{ForestGreen}{+1.91}
& \textcolor{ForestGreen}{+1.69}

\\ 
\cline{2-11}  
&\multirow{3}{*}{\makecell{DTFD-MIL \\(MaxS) } }&& 95.87&	94.54&	95.35&	96.18&
90.41&	88.40&	88.57&	94.88\\

&&+IBMIL & 96.95	&95.19&	96.12&	96.28 
&91.17&	90.89&	90.95&	96.57 \\ 
&&${\Delta}$
& \textcolor{ForestGreen}{+1.08}
& \textcolor{ForestGreen}{+0.65}
& \textcolor{ForestGreen}{+0.77}
& \textcolor{ForestGreen}{+0.10}
& \textcolor{ForestGreen}{+0.76}
& \textcolor{ForestGreen}{+2.49}
& \textcolor{ForestGreen}{+2.38}
& \textcolor{ForestGreen}{+1.69}\\ 
\hline

% #############
\multirow{12}{*}{\rotatebox{90}{\makecell{ViT \\MoCo V3}}}
&\multirow{3}{*}{ABMIL}&&89.95	&83.97&	86.82&	83.94& 88.72&	88.51&	88.57&	92.71 \\

&&+IBMIL & 87.94&	86.18&	87.60&	91.31 & 89.09&	89.02&	89.05&	93.53 \\ 
&&${\Delta}$
& \textcolor{gray}{-2.01}
& \textcolor{ForestGreen}{+2.86}
& \textcolor{ForestGreen}{+0.78}
& \textcolor{ForestGreen}{+7.37}
& \textcolor{ForestGreen}{+0.37}
& \textcolor{ForestGreen}{+0.51}
& \textcolor{ForestGreen}{+0.48}
& \textcolor{ForestGreen}{+0.82}
\\ \cline{2-11}  
&\multirow{3}{*}{DSMIL}&& 86.72&	77.24&	81.40&	82.27& 90.26&	91.37&	90.00&	95.40\\
&&+IBMIL & 85.08	&78.82&	82.17&	83.77 &91.52&	90.42&	90.48&	96.20\\ 
&&${\Delta}$
& \textcolor{gray}{-1.64}
& \textcolor{ForestGreen}{+1.58}
& \textcolor{ForestGreen}{+0.77}
& \textcolor{ForestGreen}{+1.50}
& \textcolor{ForestGreen}{+1.26}
& \textcolor{gray}{-0.95}
& \textcolor{ForestGreen}{+0.48}
& \textcolor{ForestGreen}{+0.80}
\\ \cline{2-11}  
&\multirow{3}{*}{TransMIL}&&94.21&	92.93&	93.80&	94.38 &94.26&	93.83&	93.81&	96.67 \\

&&+IBMIL & 93.84&	93.24&	93.80&	95.20 &94.34&	95.25&	94.29&	97.98\\ 
&&${\Delta}$
& \textcolor{gray}{-0.37}
& \textcolor{ForestGreen}{+0.31}
& \textcolor{gray}{0.00}
& \textcolor{ForestGreen}{+0.82}
& \textcolor{ForestGreen}{+0.08}
& \textcolor{ForestGreen}{+1.42}
& \textcolor{ForestGreen}{+0.48}
& \textcolor{ForestGreen}{+1.31}
\\ \cline{2-11}  
&\multirow{3}{*}{\makecell{DTFD-MIL \\(MaxS) } }&& 90.71&	88.44&	89.92&	90.96 & 89.45 &	90.89&	89.05&	94.95\\

&&+IBMIL & 92.65	&92.91&	93.02&	96.35 &90.34	&89.92&	90.00&	96.35\\ 
&&${\Delta}$
& \textcolor{ForestGreen}{+1.94}
& \textcolor{ForestGreen}{+4.47}
& \textcolor{ForestGreen}{+3.10}
& \textcolor{ForestGreen}{+5.39}
& \textcolor{ForestGreen}{+0.89}
& \textcolor{gray}{-0.97}
& \textcolor{ForestGreen}{+0.95}
& \textcolor{ForestGreen}{+1.40}
\\  
\bottomrule  
\end{tabular}

\label{tab:c16 tcga}
}
\end{table*}
\noindent\textbf{Dataset and evaluation protocol.}  
We conduct the experiments on two public WSI datasets, $i.e.$, Camelyon16~\cite{CAMELYON16} and TCGA-NSCLC. Camelyon16 is a dataset of H\&E stained slides for metastasis detection in breast cancer, consisting of 399 WSIs. Following~\cite{lin2022interventional}, we crop each WSI into $256\times 256$ non-overlapping patches, and remove the background region. 
There are roughly 2.8 million patches at 20$\times$ magnification in total, with about 7,200 patches per bag. 
TCGA-NSCLC includes two subtypes in lung cancer, $i.e.$, Lung Squamous Cell Carcinoma (LUSC) and Lung Adenocarcinoma(LUAD). The dataset consists of 1,054 WSIs.  We directly used the patches released by ~\cite{li2021dual}, which are about 5.2 million patches at 20$\times$ magnitude, with an average of 5000 patches for each bag.
Following the evaluation protocol of ~\cite{li2021dual}, we use 270 training images and 129 test images for Camelyon16, and 836 training images and 210 test images for TCGA-NSCLC(some corrupted slides are discarded). 
We  report the class-wise precision, recall, accuracy and area under the curve (AUC) scores.
% to evaluate the effectiveness of the methods. 
% \textcolor{red}{different seed?}

\noindent\textbf{Feature extractor.}
We adopt different network architectures with different training paradigms to thoroughly evaluate our IBMIL. \textbf{ResNet-18}~\cite{ResNet} is a widely used CNN-based model in our community, and we adopt the ImageNet pre-trained one released by PyTorch. \textbf{ViT-small}~\cite{vit} is a typical transformer-based model, which is good at modeling the long-range dependencies in the data. \textbf{CTransPath}~\cite{ctran} is a hybrid CNN and transformer architecture, customized for WSIs. We adopt the ViT pre-trained with MoCo V3~\cite{mocov3} and CTransPath pre-trained with semantically-relevant contrastive learning (SRCL), where the used data is about 15 million images from 9 datasets~\cite{ctran}.
Please refer to the Supplementary for more details.
% \noindent\textbf{Implementation Details.} To extract patch features, we adopt ImageNet pre-trained ResNet18, ImageNet pre-trained ViT by MoCo V3, and CTransPath, a hybrid CNN and ViT architecture for self-supervised training,  as feature extractors respectively. The adopted feature extractors cover representative methods, from supervised learning to self-supervised learning, ViT-based to ConvNet-based backbone. TBD %We first follow the default setting of the baseline methods to train the aggregator, and obtain the bag features. For interventional training, we set the number of confounders as xxx , and re-train the aggregator with 

\noindent\textbf{Aggregators for MIL models.} 
We build our proposed method upon 4 SOTA methods. 
\textbf{ABMIL}~\cite{AMIL} is a classic attention-based MIL, where the attention scores are predicted by a multi-layer perceptron (MLP).
% computes the attention-weighted sum of instances as the bag feature.
\textbf{DSMIL}~\cite{li2021dual}, a dual-stream framework, jointly learns an instance and a bag classifier. The highest-score instance is further used to re-calibrate other instances into a bag feature. 
\textbf{TransMIL}~\cite{shao2021transmil} is a correlated MIL framework built on transformer to explore both morphological and spatial information, where self-attention is used for bag aggregation. \textbf{DFTD-MIL}~\cite{zhang2022dtfd} proposes to virtually enlarge the number of bags by introducing the concept of pseudo-bags, resulting in a double-tier MIL framework. To align with DSMIL, we use the maximum attention score selection (MaxS) for the feature distillation strategy. For more results of DTFD-MIL (MaxMinS), please refer to the Supplementary.
% The above-mentioned methods cover competitive methods in different types of MIL, and we are expected to empower these baselines with our proposed method.

We use DSMIL’s code base for implementation and evaluation, and build other models based on their officially released codes. Since the feature extractors we use are all pre-trained, we can directly transform instances into feature vectors. For stages 2 and 3, all MIL models are optimized for 50 epochs with learning rate of 0.0001, and other settings are followed their official code. 
We  set the number of confounder $K=8$ and project dimension $l=128$ by default for all the main experiments. See Supplementary for more details.
% \subsection{Baseline Models}

\subsection{Experimental Results}
\begin{figure*}[t]
	\centering
	\begin{minipage}{1.0\linewidth}
	
	\centering
	\begin{minipage}{0.24\linewidth}
		\centering
		\includegraphics[width=0.93\linewidth]{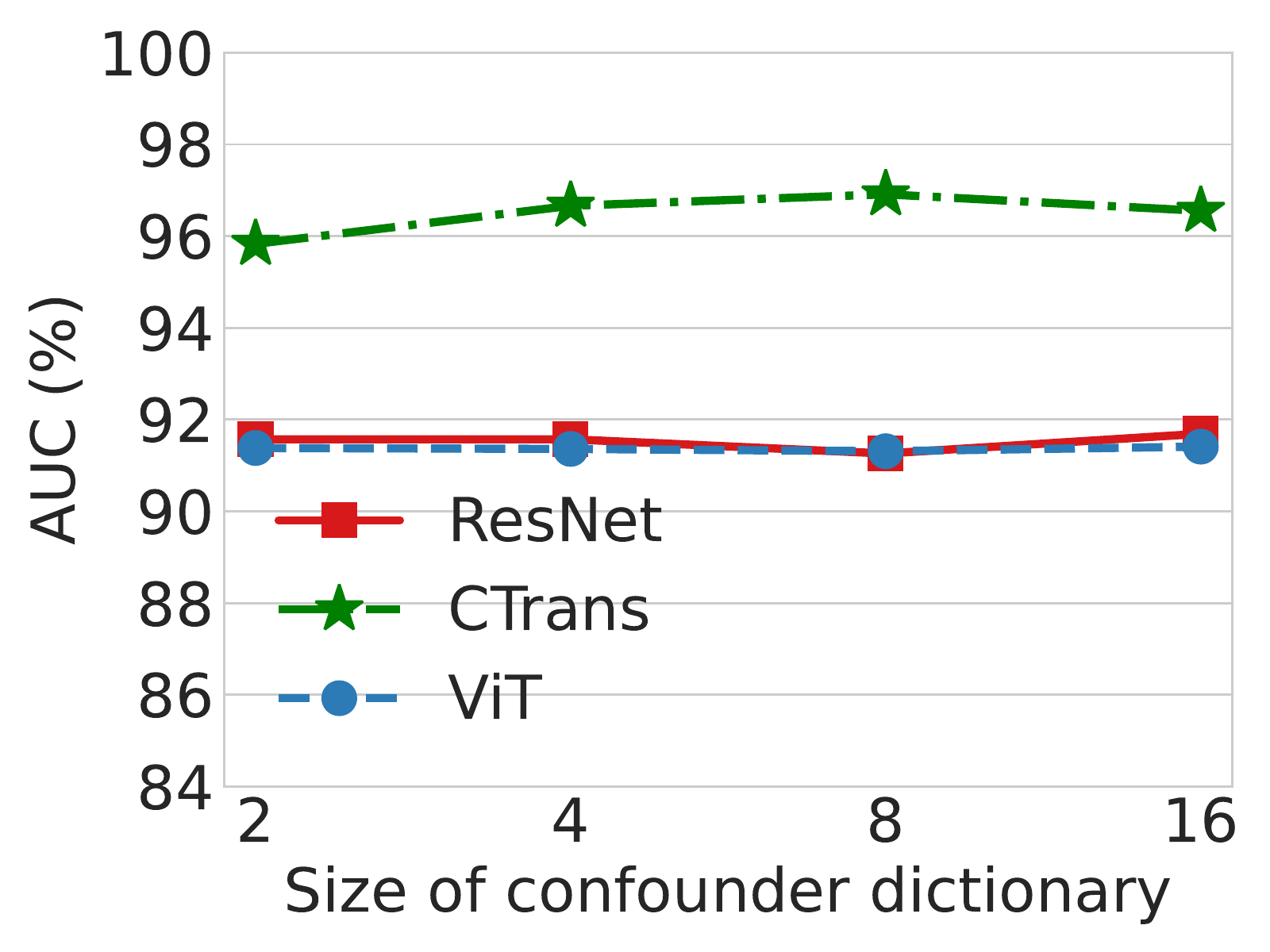}
		\subcaption{}
		\label{ab:dic size}%文中引用该图片代号
	\end{minipage}
	\begin{minipage}{0.24\linewidth}
		\centering
		\includegraphics[width=0.93\linewidth]{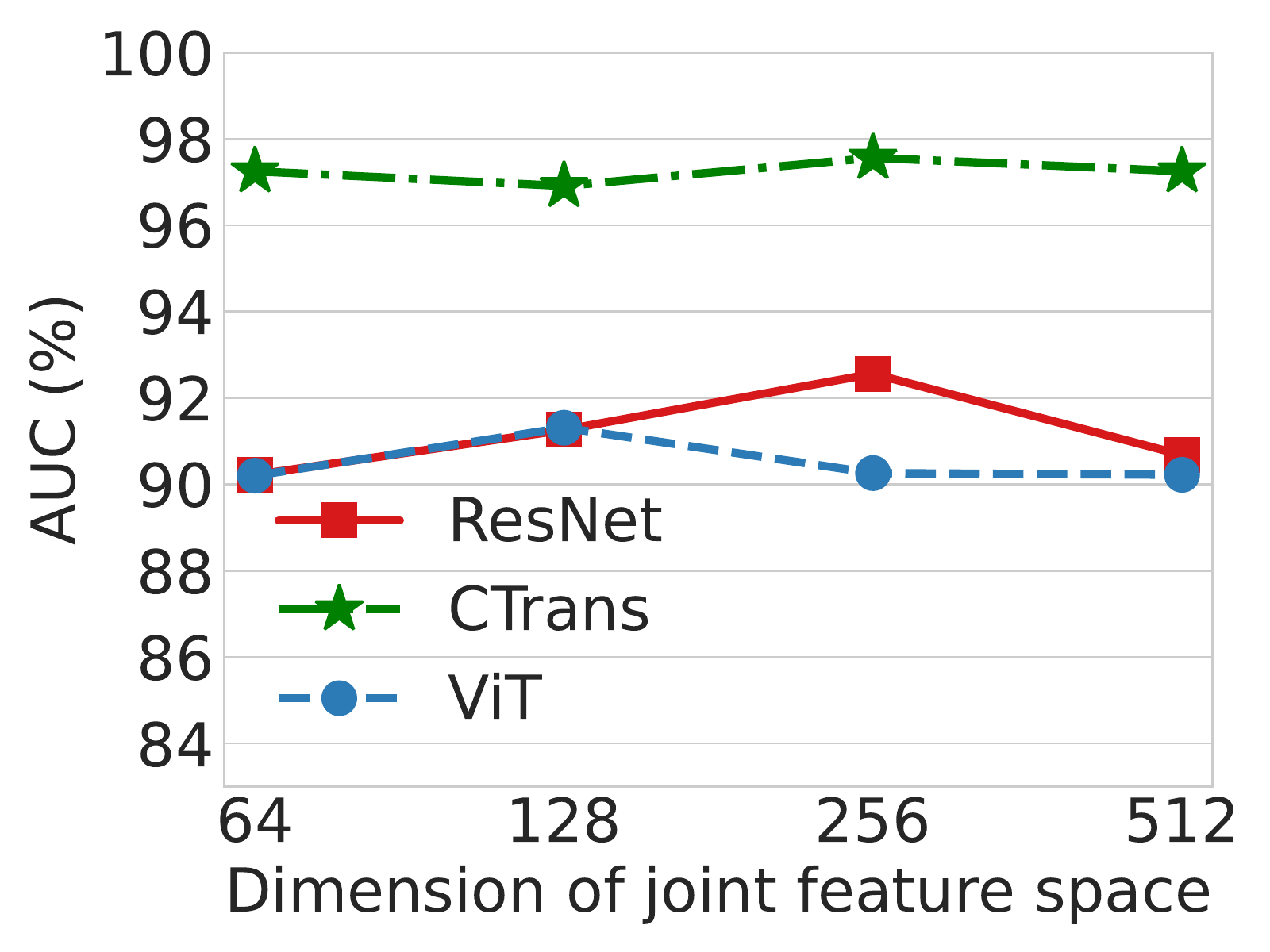}
		\subcaption{}
		\label{ab: feat dim}%文中引用该图片代号
	\end{minipage}
	\begin{minipage}{0.24\linewidth}
		\centering
		\includegraphics[width=0.93\linewidth]{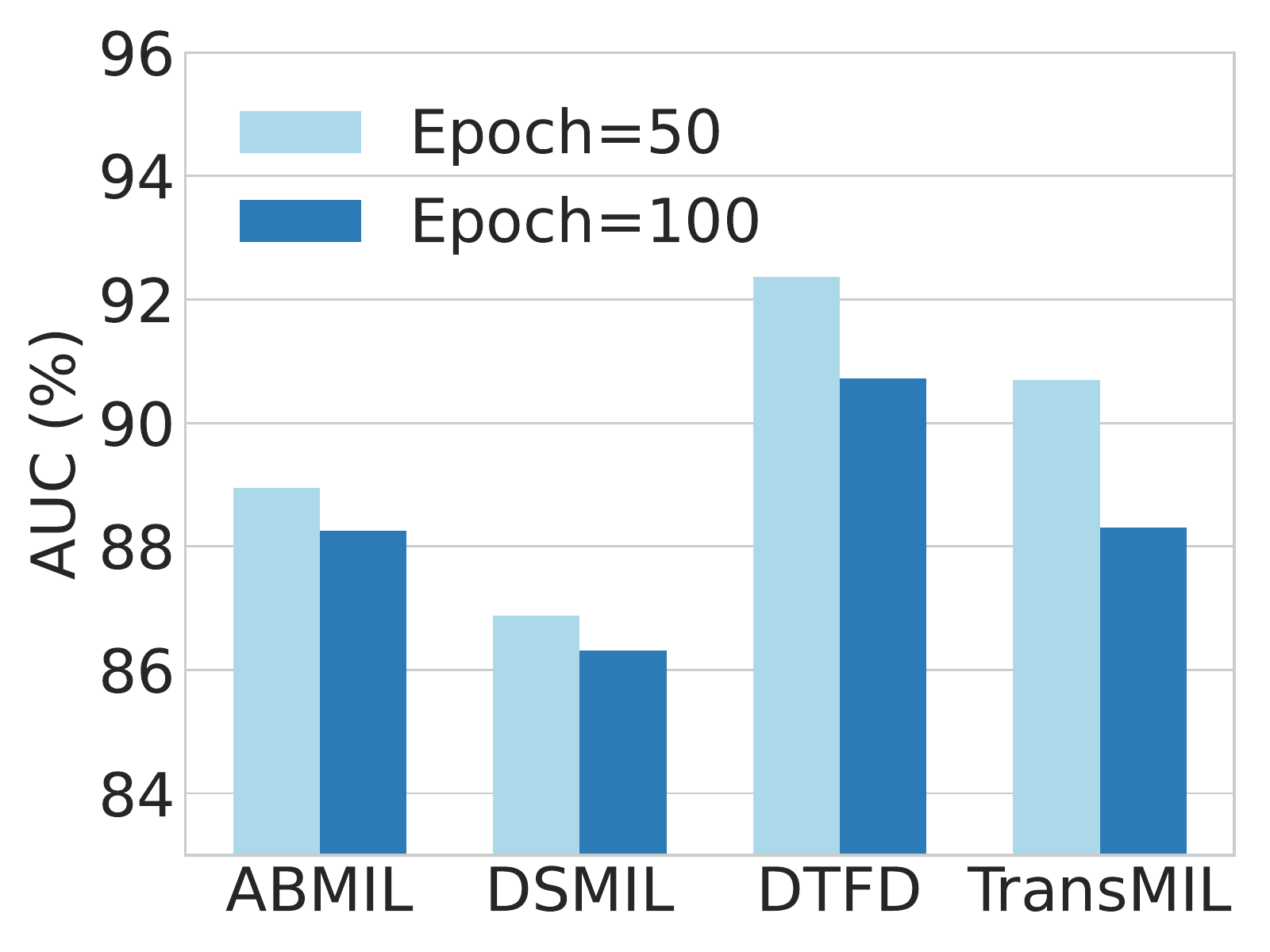}
		\subcaption{}
		\label{ab: epoch}%文中引用该图片代号
	\end{minipage}
	\begin{minipage}{0.24\linewidth}
		\centering
		\includegraphics[width=0.93\linewidth]{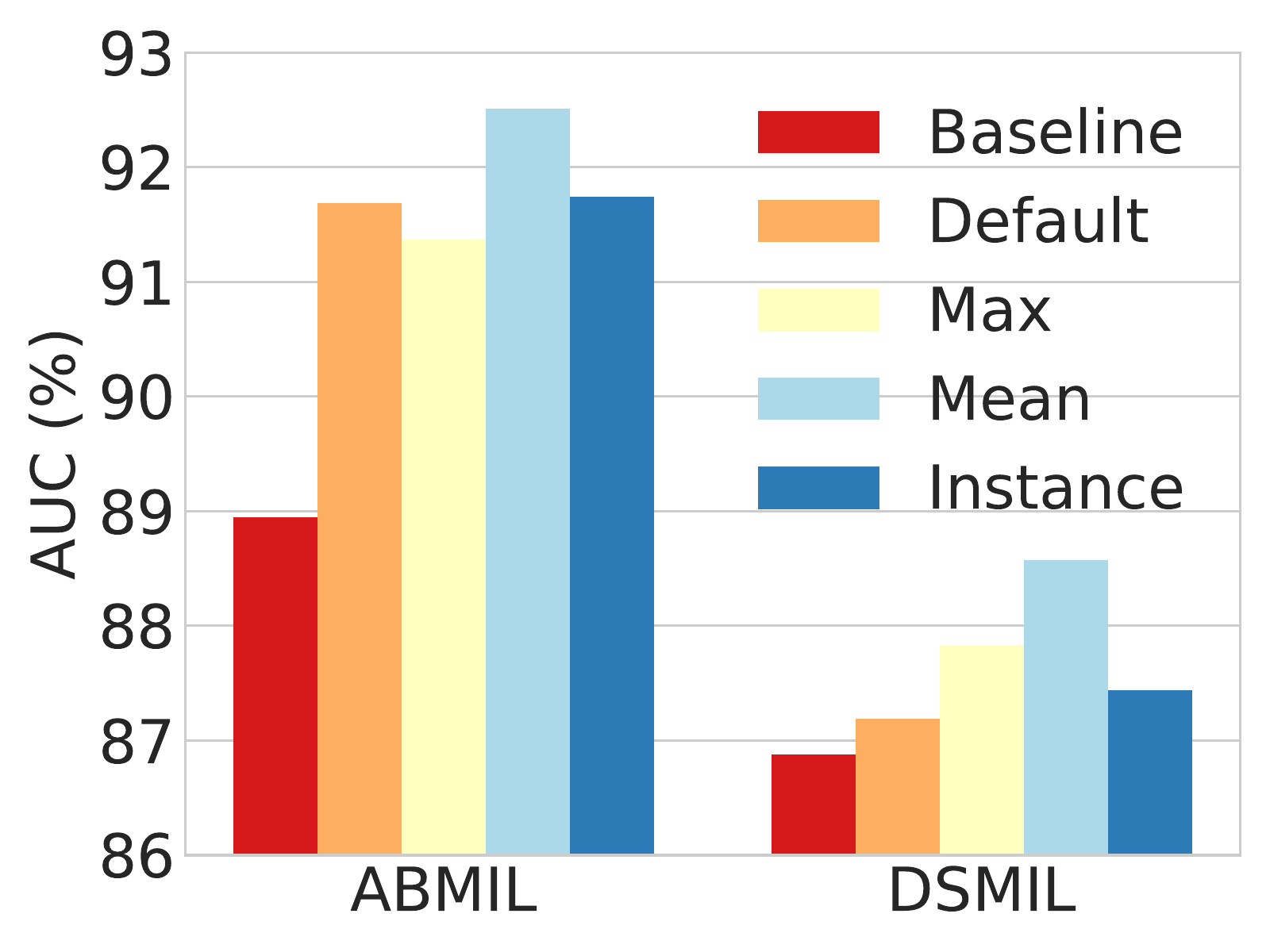}
		\subcaption{}
	\label{ab: nopara}%文中引用该图片代号
	
	\end{minipage}
	%\qquad
	%让图片换行，
	
	\end{minipage}

	\caption{Ablation studies of (a) Size of confounder dictionary, (b) Dimension of joint feature space, (c) More training epochs of baselines, and (d) Means to achieve bag features. ``Default" in (d) denotes our default 3-stage scheme.}
\end{figure*}
We present the results on two benchmark WSI datasets, Camelyon16 and TCGA-NSCLC, covering binary class MIL with unbalanced bags and multiple class MIL with balanced bags, respectively. 
By ``unbalanced'', it means only a small portion of positive instances in positive bags, $e.g.$, roughly $\textless 10\%$ in Camelyon16~\cite{li2021dual}.
From~\cref{tab:c16 tcga}, we observe that 1) IBMIL consistently improves all feature extractors with all aggregators (12 possible combinations) on both datasets, which suggests that IBMIL is agnostic to feature extractors, aggregators and datasets.
2) In particular, we find the improvement on the ImageNet pre-trained ResNet is larger than others. For example, the average gain of AUC is 5.4\% in Camelyon16 and 1.5\% in TCGA-NSCLC. This is mainly because  ResNet is more likely to learn context patterns as it is supervised trained on ImageNet~\cite{imgcontext}, while the other two are self-supervised trained with strong data augmentations --- the ``physical intervention''. 3) Our IBMIL improves more on Camelyon16 than TCGA-NSCLC in most cases. The main reason is that the former is a binary class MIL with unbalanced bags, which suffer more severe bag contextual prior --- learning the key instances is much harder than context information.
Note that the performance could be further improved by tuning the number of confounders for each setting.
% 4) Our IBMIL achieves new SOTA performance, where the AUC can be up to 97.00 
% 1. improve all. 2.resnet improve more, catch context. 3. more on c16,binary, neg ins in both bags,thus more improvement. 4. sota

% \begin{figure}[t]
%   \centering
% %   \vspace{-0.3cm}
%     \begin{subfigure}[b]{0.25\linewidth}
%     \centering
%      \includegraphics[height=3cm]{figs/num_K.pdf}
%      \caption{Causal graph}
%      \label{fig:K}
%     \end{subfigure}
%     %  \hfill
%     \begin{subfigure}[b]{0.25\linewidth}
%     \centering
%      \includegraphics[height=3cm]{figs/num_dim.pdf}
%      \caption{test}
%      \label{fig:dim}
%     \end{subfigure}
%     \caption{An illustration of causal graph for bag MIL framework.}
%     \label{fig:K_dim}
% \end{figure}

\subsection{Ablation on Model Design Variants}
\label{sec:ablation}
%We conduct extensive experiments to explore different model design choices. 
% We ablate different model design variants on TCGA-NSCLC dataset. 
In \cref{sec:ablation} and \cref{sec:dis}, experiments are conducted on TCGA-NSCLC dataset with feature extractor of ResNet-18 and aggregator of ABMIL, unless specified otherwise.
%We ablate different model design variants on TCGA-NSCLC dataset with feature extractor of ResNet-18 and aggregator of ABMIL, unless specified otherwise.
 
% \textcolor{red}{The default IBMIL is built upon ABMIL method with a feature extractor of ResNet-18,  unless specified otherwise. CAN BE REMOVED}
%and we build IBMIL upon ABMIL method. We adpot ResNet-18 as feature extractor and fix the size of dictionary to 8, unless specified otherwise.

%\input{tables/ab_K}

% zhimiao：这边打算放一些直观的模型设计消融
\noindent\textbf{Size of confounder dictionary.}
We ablate size $K$ of the confounder dictionary on three feature extractors, including ResNet-18, CTransPath and ViT. From \cref{ab:dic size}, the performance of IBMIL is relatively robust to the size of confounder dictionary. Therefore, we need not elaborately tune this hyper-parameter and an arbitrary size within a wide range is able to boost the performance. 
% For simplicity, we set $K=8$ as the default configuration.
%we set $K=8$ in subsequent ablation studies. 
%we observe no significant difference in performance with varied size of confounder dictionary. 

\noindent\textbf{Dimension of joint feature space.}
As mentioned above, the confounders and bag features are projected into a joint feature space with a dimension of $l$ and attention scores are calculated subsequently. We ablate dimension $l$ on three feature extractors. The results in \cref{ab: feat dim} reveal that performance does not improve monotonically with increased dimension and is saturated at $l=256$. We choose a dimension of 128 as the default configuration.

\noindent\textbf{Learnable vs. unlearnable confounders.}
We explore the effect of learnable and unlearnable confounders. For the former, we update them in an end-to-end manner via backpropagation.
\begin{center}
  \resizebox{0.8\linewidth}{!}{
{
\begin{tabular}{ccccc}
% \toprule
  Learnable   & Precision   & Recall   & Accuracy   & AUC  \\ \hline
\CheckmarkBold & 83.81 & 83.82 & 83.81  & 90.82 \\
\XSolidBrush  & 85.42 & 85.17 & 85.24 & 91.26  \\

% \bottomrule
\end{tabular}}}
\end{center}
As can be seen, both of them outperform the baseline accuracy of 81.43\%. However, freezing confounders during interventional training beats learnable confounders by 1.43\% on accuracy.
The reason may be that it is challenging to learn both confounders and interventional training with only bag-level labels, and introducing context-level supervision could be a solution~\cite{wang2020visual}.
We set confounders unlearnable as the default configuration.

\noindent\textbf{Implementation of backdoor adjustment.}
We study the effect of different implementations of backdoor adjustment. Given a bag feature $B \in \mathbb{R}^d$ and the combination of confounders $\sum\nolimits_{i=1}^K \alpha_i {c}_i P(c_i) \in \mathbb{R}^d$, we explore three variants to combine them, $i.e.$ $B \star \sum\nolimits_{i=1}^K \alpha_i {c}_i P(c_i)$ and $\star \in \{\oplus ,+, - \}$, where $+/-$ is element-wise addition/subtraction.
\begin{center}
{\resizebox{0.8\linewidth}{!}{
\begin{tabular}{cccccc}
% \toprule
Method                    & Precision   & Recall   & Accuracy   & AUC  \\ \hline
$\oplus$  & 85.42 & 85.17 & 85.24 & 91.26 \\
 $+    $                        & 84.99 & 84.68 & 84.76 & 89.28 \\
 $-$                            &84.70 & 84.18 &84.29 & 90.14 \\

% \bottomrule
\end{tabular}}}
\end{center}
We observe all these implementations can lead to performance improvements, which demonstrates the stability  and effectiveness of the proposed intervention. 

% \input{tables/ab_imple}

%\textcolor{red}{May mention necessity of stage 2 here.}
%Noticing our proposed method needs an extra step for training aggregators, a natural question arises: Does longer training time of baseline methods improve the performance ? As a response, we prolong the training epochs of baselines from 50 to 100. Table xxx reports the results and no obvious performance improvement is observed.

%\input{figs/total_ab_part2}
\subsection{Analysis and Discussion.}
\label{sec:dis}

% \cref{tab:ana_cls} reports the performance of different choices of confounder set constitution. For class-specific constitution method, We observe significant performance degradations on Camelyon16 dataset with ViT compared to its counterpart, while other results remain comparable.
% Experiments indicate that 
% The appropriate confounder approximations can be obtained on unlabelled datasets.
%In actual fact, confounders could be independent of the class identity
%We hypothesize that class-specific constitution is not suitable for IBMIL when the dataset is not very large or the feature extractor is not very powerful. Besides, the confounder could be independent of the class identity.

\noindent\textbf{Do improvements come from more epochs?}
%Compared with baseline methods, our proposed method needs an extra training stage . 
Note  that our proposed method requires an extra stage to train the aggregator. A natural question is whether we can improve baseline performance by training the baseline methods for as many epochs as the extra stage. \cref{ab: epoch} displays that  more epochs do not bring about performance improvement, showing that our proposed method could empower baseline methods by backdoor adjustment rather than more training epochs. In most cases, training longer even brings performance degradation, which can be caused by the over-fitting problem in MIL~\cite{zhang2022dtfd}.

% Besides, further explorations in next paragraph demonstrate that the extra training stage can be discarded without affecting performance in most cases.
% \input{tables/ana_cls}
% \input{tables/ab_time}

\begin{figure*}[t]
	\centering
	\begin{minipage}{1.0\linewidth}
	
	\centering
	\begin{minipage}{0.19\linewidth}
		\centering
		\includegraphics[width=0.98\linewidth]{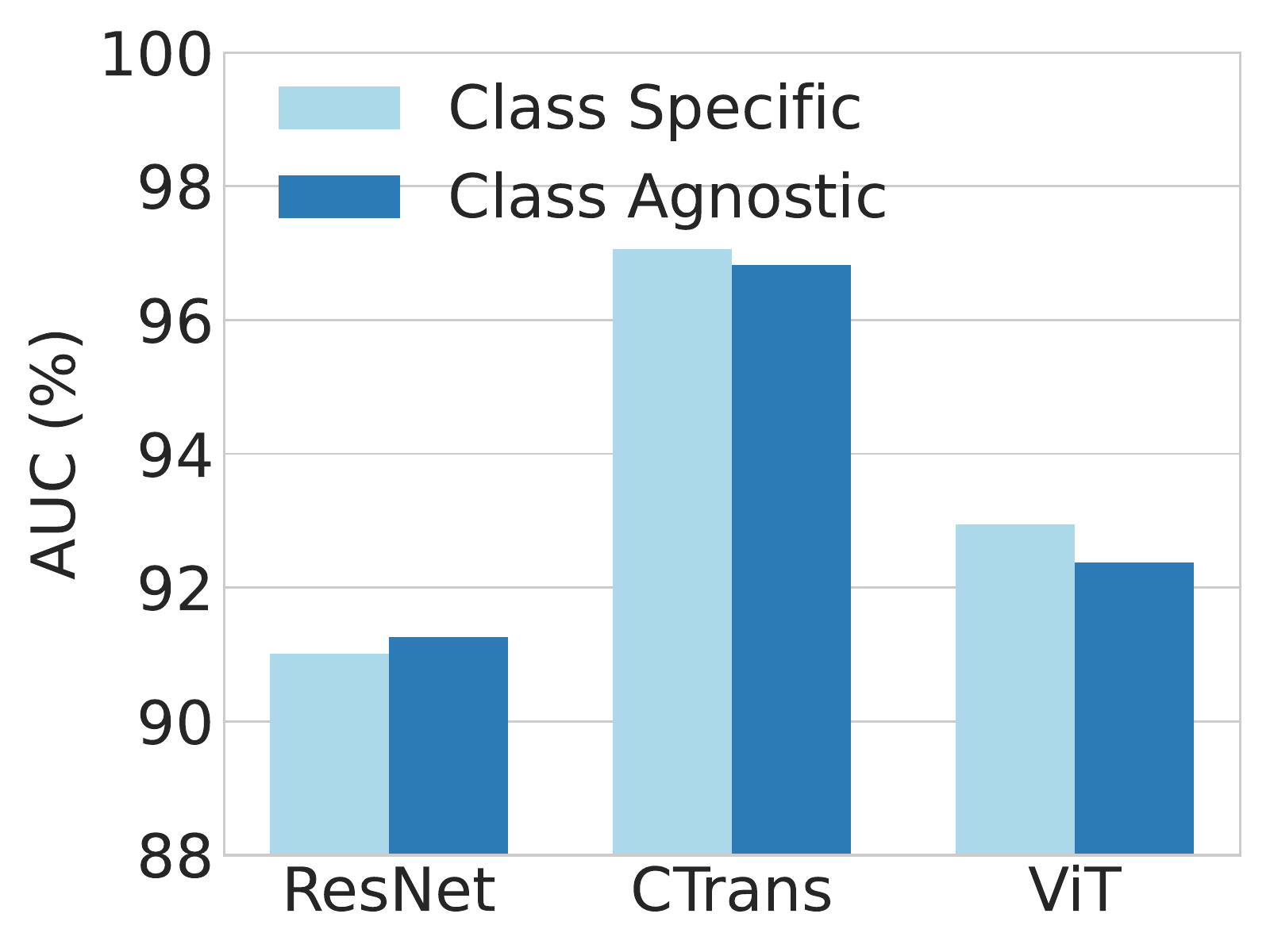}
		\subcaption{}
		\label{ab:cls_spec}%文中引用该图片代号
	\end{minipage}
	\begin{minipage}{0.19\linewidth}
		\centering
		\includegraphics[width=0.98\linewidth]{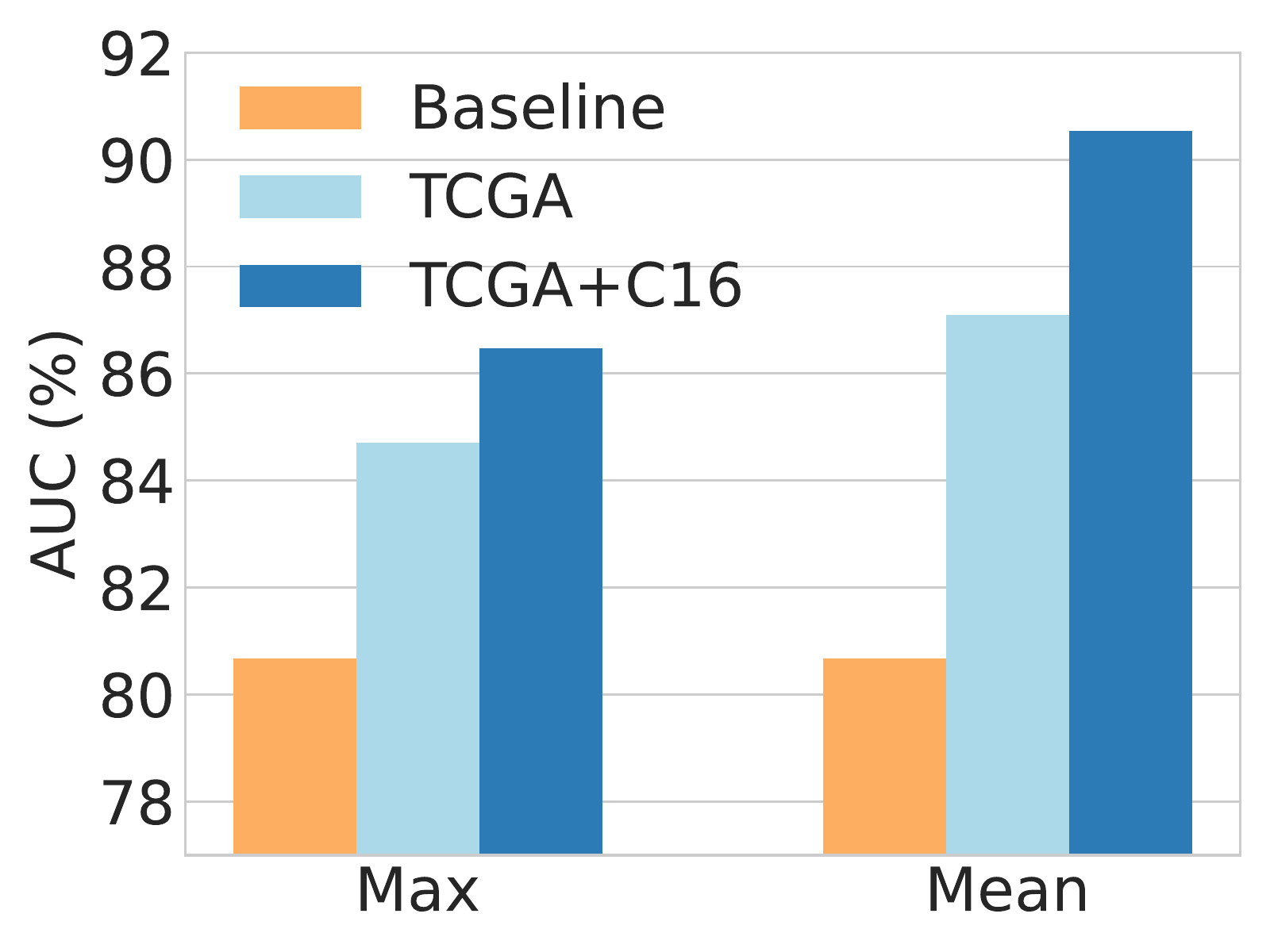}
		\subcaption{}
		\label{ab:merge}%文中引用该图片代号
	\end{minipage}
	\begin{minipage}{0.19\linewidth}
		\centering
		\includegraphics[width=0.98\linewidth]{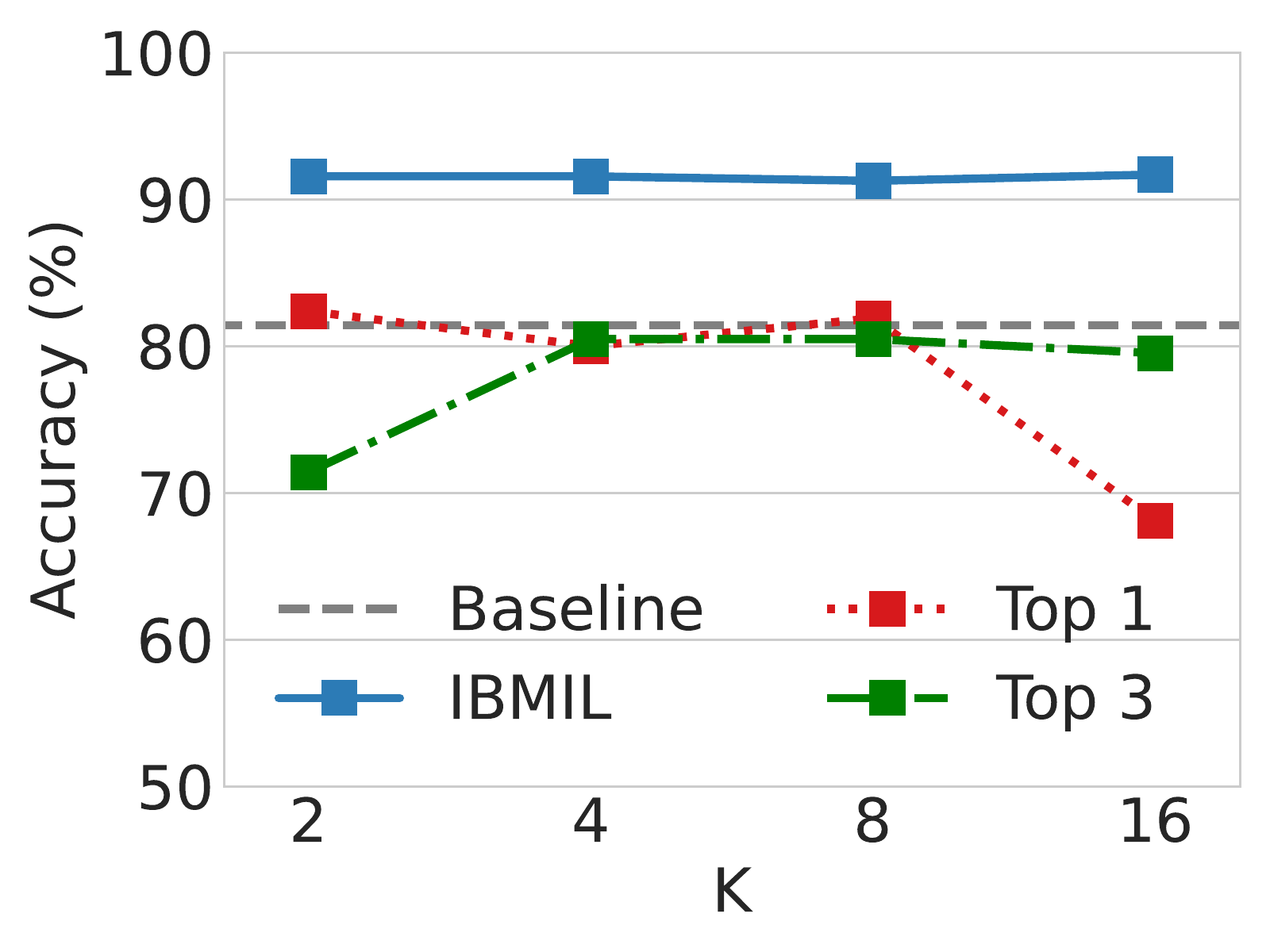}
		\subcaption{}
		\label{ab:resnet}%文中引用该图片代号
	\end{minipage}
	\begin{minipage}{0.19\linewidth}
		\centering
		\includegraphics[width=0.98\linewidth]{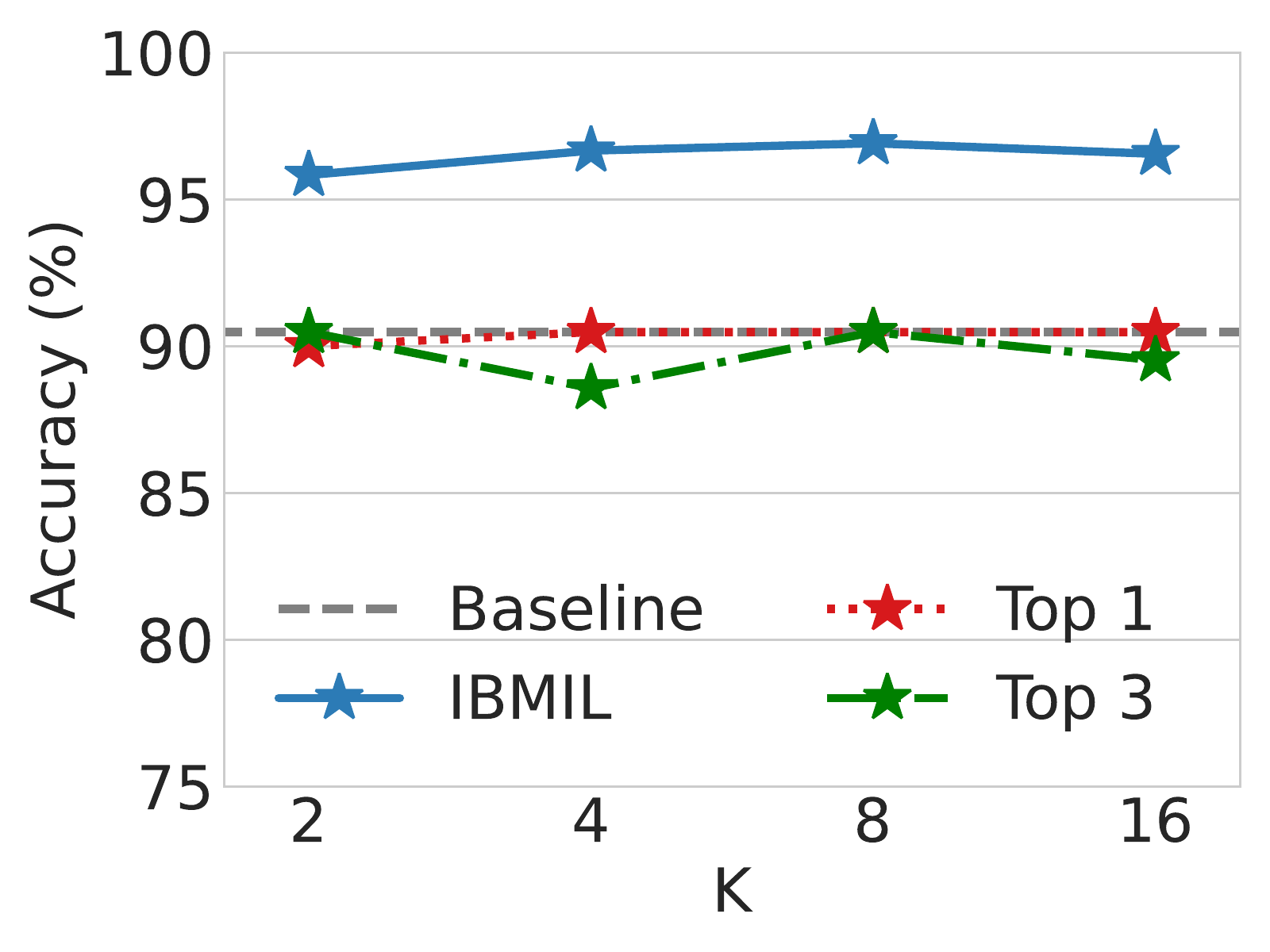}
		\subcaption{}
		\label{ab:ctrans}%文中引用该图片代号
	\end{minipage}
	\begin{minipage}{0.19\linewidth}
		\centering
		\includegraphics[width=0.98\linewidth]{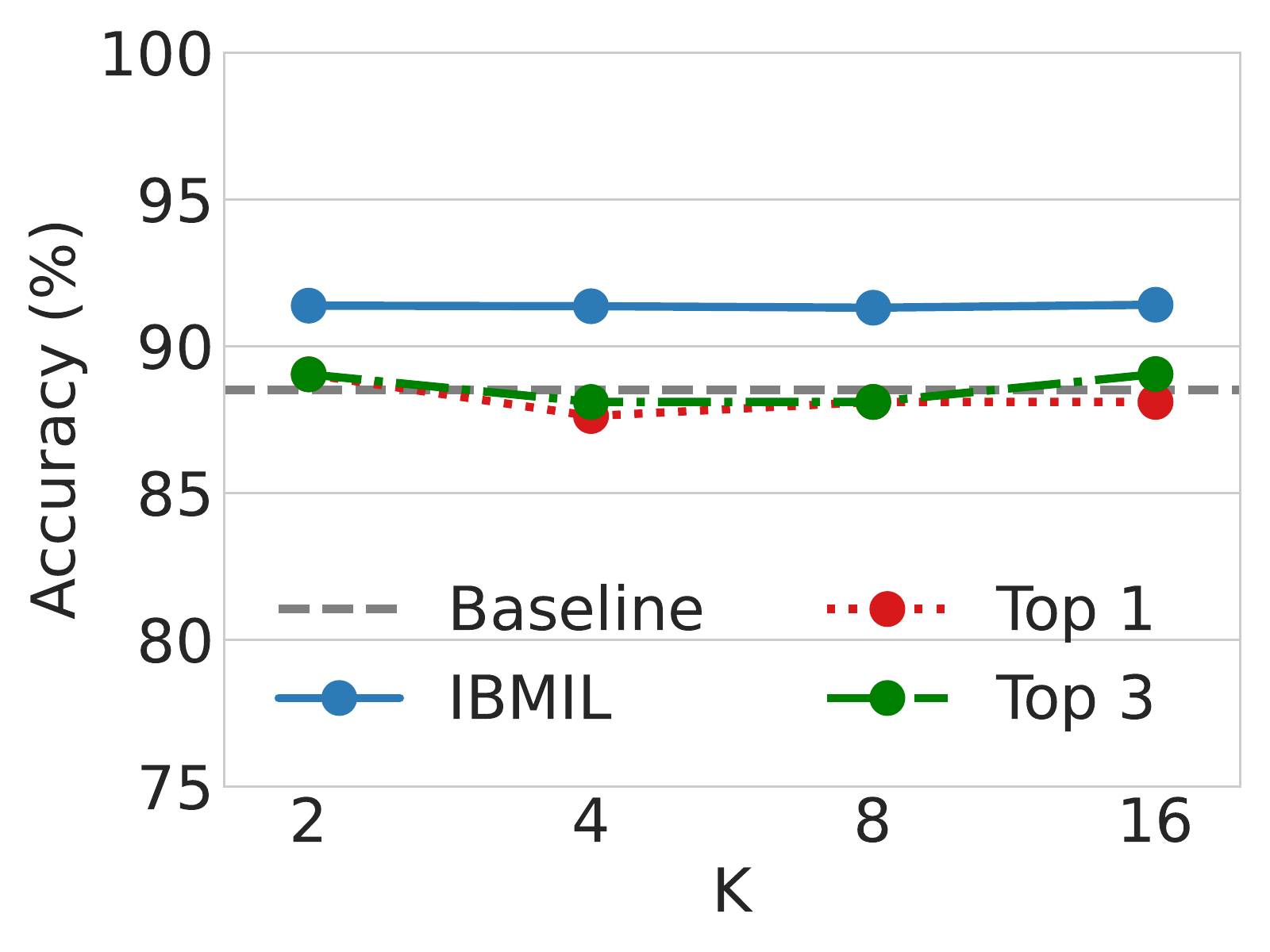}
		\subcaption{}
		\label{ab:vit}%文中引用该图片代号
	\end{minipage}
	%\qquad
	%让图片换行，
	
	\end{minipage}

	\caption{(a) Constitution of confounder set: class-specific vs. class-agnostic. (b) Confounder set from TCGA vs. TCGA+Camelyon16. (c)-(e) Intervention training vs. KNN classifier with the feature extractors of ResNet, CTransPath and ViT, respectively. }
\end{figure*}
\noindent\textbf{Is stage 2 necessary?}
Recent MIL methods aggregate the instance features into a bag feature via the weighted average operator. The weights, also referred to as attention scores, are generated by parametric networks, which need an extra stage of training aggregators.
Alternatively, we turn to three non-parametric settings to skip this stage and efficiently achieve the bag features. We consider:
\begin{itemize}
    \item ``Mean" / ``Max" denotes a bag feature is obtained through a mean-pooling / max-pooling layer among a bag of instance features, which is inspired by the strong baseline of non-parametric MIL method~\cite{wang2018revisiting}.
    %\item ``Max" denotes a bag feature is obtained through a Max-Pooling layer among a bag of instance features, and $K$-means is performed over all the bag features subsequently.
    \item ``Instance" denotes that $K$-means is directly performed over all the instance features in training set, since each instance can be regarded as a bag with length of one.
\end{itemize}
Then, interventional training is applied to baseline methods (including ABMIL and DSMIL) and we report the results in \cref{ab: nopara}. 
% ``/" and  ``Default" denote the baseline and the standard three-stage scheme, respectively. 
Notably, even with such simple aggregation strategy, IBMIL still outperforms the baseline, and remains competitive or even better compared to  ``Default" setting, which indicates stage 2 in our scheme is unnecessary. By omitting stage 2, our scheme can be more elegant without performance degradation in most cases.

% % Please add the following required packages to your document preamble:
% % \usepackage{multirow}
\begin{table}[t]
\caption{Performance of non-parametric aggregators.}
\label{tab:ana_simple}
\centering
\resizebox{0.95\linewidth}{!}{
\begin{tabular}{cccccc}
\toprule
Aggregator                   & $K$        & Precision   & Recall   & Accuracy   & AUC  \\ \hline
\multirow{5}{*}{Max}  & /  & \color{gray}{77.59} &	\color{gray}{77.35} &	\color{gray}{70.95} &	\color{gray}{82.23} \\
                      & 2  & 81.58 &	73.54 &	72.38 &	83.44 \\
                      & 4        & 78.34&	79.17&	76.67&	84.71 \\
                      & 8        & 80.41 &	78.72 &	78.10 &	84.95 \\
                      & 16        & 82.32  &	71.06&	70.48&	83.79 \\
\midrule

\multirow{4}{*}{Mean}  & /  & \color{gray}{77.13}	& \color{gray}{71.60}	&\color{gray}{71.43} & \color{gray}{80.68} \\
                      & 2        & 81.97 &	81.49 &	81.43 &	85.81 \\
                      & 4        &82.04	& 81.56	& 81.43	 & 87.10 \\
                      & 8        &84.50 &	80.46 &	80.48 &	89.14 \\
                      & 16        & 85.13 &	78.16 &	78.10 &	89.00 \\

\bottomrule
\end{tabular}
}
\end{table}

% \usepackage{multirow}
% \begin{table*}[t]
% \caption{Performance of non-parametric aggregators.}
% \label{tab:ana_simple}
% \centering
% \resizebox{0.9\linewidth}{!}{
% \begin{tabular}{cccccccccccc}
% \toprule
% Method    & $K$        & Precision   & Recall   & Accuracy   & AUC &Method    & $K$        & Precision   & Recall   & Accuracy   & AUC \\ \hline
% \multirow{5}{*}{\rotatebox{90}{Max}}  & /  & \color{gray}{77.59} &	\color{gray}{77.35} &	\color{gray}{70.95} &	\color{gray}{82.23} & \multirow{5}{*}{\rotatebox{90}{Mean}}  & /  & \color{gray}{77.13}	& \color{gray}{71.60}	&\color{gray}{71.43} & \color{gray}{80.68} \\
% & 2  & 81.58 &	73.54 &	72.38 &	83.44 & & 2        & 81.97 &	81.49 &	81.43 &	85.81\\
% & 4        & 78.34&	79.17&	76.67&	84.71 && 4        &82.04	& 81.56	& 81.43	 & 87.10\\
% & 8        & 80.41 &	78.72 &	78.10 &	84.95 && 8        &84.50 &	80.46 &	80.48 &	89.14\\
% & 16        & 82.32  &	71.06&	70.48&	83.79& & 16        & 85.13 &	78.16 &	78.10 &	89.00 \\
% \bottomrule
% \end{tabular}
% }
% \end{table*}

\noindent\textbf{Can IBMIL improve non-parametric baselines?}
Besides using non-parametric aggregators to generate bag features for confounder set, we further take them as baselines and verify whether IBMIL is also able to improve them ($i.e.$, max/mean-pooling).
Surprisingly, in~\cref{tab:ana_simple}, IBMIL brings significant improvements under all settings, where the best performance is even comparable to these attention-based aggregators.
It indicates that IBMIL is indeed compatible with all compared MIL methods, including the non-parametric ones.
% To verify this, we , and report the performance . Unsurprisingly, the baselines (denoted as ``/") \textcolor{red}{LTCCCCCCCCCCCC}

%\input{tables/ana_cls}
\noindent\textbf{Constituting  confounder set w/  or w/o bag labels?}
% \textcolor{red}{Constitution of confounder set or dictionary ?}
% Our default IBMIL constitutes the confounder set without using the bag labels, $i.e$, in a class-agnostic manner. 
% From another view, we can divide bag features into different groups according to their labels, and $K$-means is performed on each separate group subsequently. In this way, the intra-class variation is preserved, and the confounder dictionary captures the class-relevant chawe exploreracteristics.
Given bag labels, we explore the class-specific  $K$-means. In particular,
we  apply $K$-means to each class respectively, preserving the intra-class variation and class-relevant characteristics.
From~\cref{ab:cls_spec}, we observe no obvious performance gap between class-specific and class-agnostic $K$-means. 
We conjecture that 1) the confounders could be independent of the class identity, and 2) bag features are already separable by classes. We will explore the way of incorporating bag labels in future work.
On the other hand, the unsupervised fashion makes our scheme compatible with large-scale unlabelled datasets. We explore more unlabelled bags via  combining the bags of TCGA and Camelyon16, and constituting the confounder set via the non-parametric aggregators.
From~\cref{ab:merge}, we observe a clear improvement on AUC under both max- and mean-poolings.
That indicates, with more bags, our implementation can achieve better approximation of confounders.

\noindent\textbf{Is IBMIL just post-processing?}
% zhimiao：这边讨论对Confounder的设计，有意义的、深入的分析 
Since our proposed IBMIL shares some commonalities with the embedding-based MIL~\cite{wang2018revisiting}, one may ask: Do the improvements only come from exploring the bag relations?
To answer this question, we make minor modifications on ABMIL. Instead of interventional training with confounders, we obtain the confounder dictionary via the class-specific  $K$-means and treat it as a KNN classifier for evaluation. 
% (See Supplementary for more details.)
%infer via confounder set using the KNN classifier. 
As can be seen, it  brings limited improvements or even degrades the performance, verifying the improvements comes from interventional training, which is not just post-processing.

\section{Conclusions}
The vast majority of recent efforts in this field seek to enhance the feature extractor and aggregator. This paper addresses MIL from a novel perspective via analyzing the confounders between bags and labels. This leads to the proposed novel Interventional Bag Multi-Instance Learning (IBMIL), a new deconfounded bag-level prediction approach to suppress the bias caused by the bag contextual prior. IBMIL introduces a structure causal model to reveal the causalities and eliminates their effect through the backdoor adjustment with practical implementations. Comprehensive experiments have been conducted on various MIL benchmarks and the results show that IBMIL can boost existing methods significantly. In future, we plan to approximate confounder set in a more efficient and elegant manner. As a general method to use causal intervention for bag-level prediction, IBMIL provides fresh insight into MIL problem.

\noindent\textbf{Acknowledgments.}  
Dr. Yi Xu was supported in part by NSFC 62171282, Shanghai Municipal Science and Technology Major Project (2021SHZDZX0102), 111 project BP0719010, and SJTU Science and Technology Innovation Special Fund ZH2018ZDA17 and YG2022QN037. 

%%%%% is that%%%% REFERENCES

%%%% method%%%%% REFERENCES
{\small
\bibliographystyle{ieee_fullname}
\bibliography{egbib}
}

\end{document}